\documentclass[11pt]{article}

\usepackage[final]{acl}

\usepackage{times}
\usepackage{latexsym}

\usepackage[T1]{fontenc}

\usepackage[utf8]{inputenc}

\usepackage{microtype}

\usepackage{inconsolata}

\usepackage{graphicx}

\usepackage{subcaption}
\usepackage{float} 

\usepackage{soul}
\usepackage{xcolor}
\usepackage{booktabs}
\usepackage{array}
\usepackage{tabularx}
\usepackage{amsmath}
\usepackage{multirow}

\definecolor{actionbg}{RGB}{250, 238, 218}
\definecolor{actionfg}{RGB}{133, 79, 11}
\definecolor{perceivedbg}{RGB}{230, 241, 251}
\definecolor{perceivedfg}{RGB}{12, 68, 124}
\definecolor{descbg}{RGB}{252, 235, 235}
\definecolor{descfg}{RGB}{120, 31, 31}
\definecolor{visualbg}{RGB}{220, 245, 220}
\definecolor{visualfg}{RGB}{30, 100, 30}
\definecolor{nospacebg}{HTML}{F0F0F0}
\definecolor{nospacefg}{HTML}{4D4D4D}

\newcommand{\spacea}[1]{{\sethlcolor{actionbg}\hl{#1}}}
\newcommand{\spacep}[1]{{\sethlcolor{perceivedbg}\hl{#1}}}
\newcommand{\spaced}[1]{{\sethlcolor{descbg}\hl{#1}}}
\newcommand{\spacev}[1]{{\sethlcolor{visualbg}\hl{#1}}}
\newcommand{\spacen}[1]{{\sethlcolor{nospacebg}\hl{#1}}}

\newcommand{\abadge}{\,\raisebox{0.4ex}{\scalebox{0.65}{%
  \colorbox{actionbg}{\textcolor{actionfg}{\textsf{action space}}}}}}
\newcommand{\pbadge}{\,\raisebox{0.4ex}{\scalebox{0.65}{%
  \colorbox{perceivedbg}{\textcolor{perceivedfg}{\textsf{perceived space}}}}}}
\newcommand{\dbadge}{\,\raisebox{0.4ex}{\scalebox{0.65}{%
  \colorbox{descbg}{\textcolor{descfg}{\textsf{descriptive space}}}}}}

\newcommand{\vbadge}{\,\raisebox{0.4ex}{\scalebox{0.65}{%
  \colorbox{visualbg}{\textcolor{visualfg}{\textsf{visual space}}}}}}

\newcommand{\nbadge}{\,\raisebox{0.4ex}{\scalebox{0.65}{%
  \colorbox{nospacebg}{\textcolor{nospacefg}{\textsf{no space}}}}}}

\title{How LLMs Build Fictional Worlds: Setting and Narrative Space in AI-Generated Creative Storytelling}

\author{
\textbf{Katrin Rohrbacher}\textsuperscript{1},
\textbf{Björn Nieth}\textsuperscript{2,5},
\textbf{Emmanuelle Salin}\textsuperscript{2}, \\
\textbf{Bjoern Eskofier}\textsuperscript{2,3,5,6},
\textbf{Michaela Mahlberg}\textsuperscript{1,4}\\[0.3em]
\small \textsuperscript{1}The Text and Language Lab, Department of Digital Humanities and Social Studies (DHSS), FAU Erlangen-Nürnberg, Germany\\
\small \textsuperscript{2}Department Artificial Intelligence in Biomedical Engineering (AIBE), FAU Erlangen-Nürnberg, Germany\\
\small \textsuperscript{3}Chair of AI-supported Therapy Decisions, LMU München, Germany\\
\small \textsuperscript{4}Department of Linguistics and Communication, University of Birmingham, United Kingdom\\
\small \textsuperscript{5}Munich Center for Machine Learning (MCML), Munich, Germany\\
\small \textsuperscript{6}Institute of AI for Health, Helmholtz Zentrum München, Neuherberg, Germany\\[0.2em]
\small \texttt{\{katrin.rohrbacher, bjoern.nieth, emmanuelle.salin, bjoern.eskofier, michaela.mahlberg\}@fau.de}
}

\begin{document}
\maketitle
\begin{abstract}
In this paper, we analyze how Large Language Models (LLMs) employ worldbuilding strategies, focusing on setting as one measurable dimension of storyworld construction. We compare 1,000 AI-generated stories per model in English and German with human-authored fiction from Project Gutenberg. Building on prior work, we operationalize setting through five types of narrative space: “action,” “perceived,” “visual,” “descriptive” and “no space”, identified using fine-tuned BERT classifiers for German and English. We generate narratives using GPT~4.1, LlaMA 3.3, Mistral~3.2, and Gemma~3 and compare their spatial distributions to a human-authored baseline. We find that human-authored texts predominantly employ “action space,” grounding narratives in embodied character–environment interaction, whereas LLMs systematically overproduce “perceived space,” emphasizing atmosphere and affect. This divergence remains stable across narrative time. Overall, our findings show that LLMs exhibit worldbuilding patterns that differ consistently from human-authored fiction in ways that are both model-specific and language-sensitive.
\end{abstract}

\section{Introduction}

Research in narratology, literary studies, and linguistics has identified worldbuilding as a central element of storytelling and one of its core communicative functions. 
Herman argues that storyworlds are built from verbal and visual cues that prompt readers to mentally construct the depicted world as it evolves through interactions among characters, objects, and settings \cite{herman_story_2002}. 
In cognitive linguistics and corpus stylistics, researchers have explored how readers construct mental representations of fictional worlds drawing on ``textual building blocks'' or ``world-building elements'' \cite{mahlberg_corpus_2013, gavins_text_nodate}. Gavins, in her work on ``world-building'' in fiction, explicitly focuses on the opening paragraphs of stories, which serve as ``an initial introduction to the fictional worlds about to be realized by the text'' \citep[p.~133]{gavins_text_nodate}. These approaches emphasize that fictional worlds emerge through specific textual structures, which collectively shape how readers experience and navigate storyworlds. 

\begin{figure}
\begin{minipage}[t]{0.48\columnwidth}
\raggedright\small
\textit{Human-authored}\\[4pt]
\vspace{3pt}
\spacea{\textbf{Lilian Boyd entered the small, rather shabby room, neat, though everything was well worn.}}\abadge{}
\spaced{Her mother sat by a little work table busy with some muslin sewing and she looked up with a weary smile.}\dbadge{}
\spacea{Lilian laid a five-dollar bill on the table.}\abadge{}
\spacen{“Madame Lupton sails on Saturday,” she said.}\nbadge{} 
\spacen{“Oh how splendid it must be to go to Paris!”}\nbadge{}
\end{minipage}
\hfill\vrule width 0.4pt\hfill
\begin{minipage}[t]{0.48\columnwidth}
\raggedright\small
\textit{GPT~4.1}\\[4pt]
\vspace{3pt}
\spacea{\textbf{Lilian Boyd entered the small, rather shabby room, neat, though everything was well worn.}}\abadge{}
\spacep{A stuffy scent, reminiscent of dust and distant rain, lingered in the corners.}\pbadge{} 
[\dots]
\spacev{Lilian closed the door behind her and allowed her eyes to sweep across the sparse furnishings.}\vbadge{}
\spaced{The same old brown armchair squatted in the corner by the window.}\dbadge{}
\end{minipage}
\vspace{5pt}
\hrule
\vspace{3pt}
{\footnotesize\setlength{\fboxsep}{1.5pt}%
  \textbf{Legend:}\hspace{2.5pt}%
  \colorbox{actionbg}{\textcolor{actionfg}{action}}\hspace{2.5pt}%
  \colorbox{perceivedbg}{\textcolor{perceivedfg}{perceived}}\hspace{2.5pt}%
  \colorbox{descbg}{\textcolor{descfg}{descriptive}}\hspace{2.5pt}%
  \colorbox{visualbg}{\textcolor{visualfg}{visual}}\hspace{2.5pt}%
  \colorbox{nospacebg}{\textcolor{nospacefg}{no space}}%
}

\caption{The same opening sentence (bold) continued by a
human author (Amanda M. Douglas, \emph{The Girls at Mount Morris}, 1914) and GPT~4.1, annotated by the setting classifier.}
\label{tab:qualitative}
\end{figure}

The task of creative text generation has received increasing attention in current LLM research. Existing studies on LLMs’ capacities to generate fiction have primarily analyzed outputs through creativity tests or human judgments of stylistic quality \cite{zhao_assessing_2024, ismayilzada_evaluating_2025}.
Other studies have examined specific narrative and stylistic features, such as embodied language, emotional expression, temporal structure, and plot construction, to better understand how LLMs generate narratives \cite{hicke_zero_2025, ishikawa_ai_2025, zhong_exploring_2024, fatemi2024test, ahuja2025finding}. However, the spatial dimension of worldbuilding in AI-generated fiction remains largely unexplored.

In this study, we focus on setting as a central dimension of worldbuilding in creative storytelling, understanding it as one of the primary textual mechanisms (alongside time and events) through which fictional worlds are constructed. We build on the framework introduced in \citet{rohrbacher_opening_2025a}, which grounds narrative setting in the phenomenological notion of \textit{lived space} \cite{hoffmann1980raum, stroker_philosophische_1965}, i.e., space as it is experienced and inhabited by a perceiving subject, not as a neutral, abstract extension. Lived space, in this sense, is not simply described but organized around characters' perception and embodied engagement with their surroundings, conveying ``what it is like'' to inhabit a fictional world \cite{herman_basic_2009}. The framework defines five categories. Action space, perceived space, and visual space reflect distinct modes of spatial experience. Descriptive space, by contrast, situates characters and objects in space without being anchored to any character's point of view or experiential engagement. ``No space'' covers sentences without spatial reference. 

To assess how LLMs use these spatial categories to construct narrative worlds, we apply the setting classifier introduced in \citet{rohrbacher_opening_2025a}, a fine-tuned BERT model originally developed for German fiction and extended in this study to English. We generate 1,000 stories per model and language and compare the resulting distributions against a human-authored corpus serving as a baseline.

The contributions of this paper are as follows:

\begin{itemize}

\item We find that LLM storytelling is systematically more atmospheric and less embodied/action-driven than human fiction. 

\item Through a cross-linguistic comparison of English and German corpora, we show that these deviation patterns are jointly shaped by model and language context.

\item We release a large-scale dataset of 8,000 AI-generated stories (1,000 per model across four LLMs and two languages), matched with a human-authored corpus. 

\item We release an English setting classifier and accompanying annotation resources that replicate an existing German classifier, supporting corpus-scale, narratology-informed evaluation of AI-generated fiction. 

\end{itemize}

\section{Related Work} 

Despite the extensive narratological literature on how stories are told, relatively little machine learning research has drawn on narratological concepts to analyze how LLMs generate narratives. Existing approaches to evaluating LLM-generated narratives include creativity tests \cite{chakrabarty_art_2024, ismayilzada_evaluating_2025}, often adapted from psychological frameworks such as the Torrance Test of Creative Thinking \cite{zhao_assessing_2024, marco_pron_2024, chen_probing_2023}, which primarily measure human creative cognition rather than how creativity is realized in textual form. Other work has used human raters, for instance to assess stylistic properties in order to infer perceived quality or humanlikeness \cite{chakrabarty_readers_2025}, or to identify a story's origin and rate its quality \citep{sears_bot_2026}. A third line of work focuses on identifying specific storytelling patterns in generated prose, including lexicon- or dictionary-based approaches for capturing concepts such as embodiment in narrative \cite{hicke_zero_2025}. However, such approaches struggle with polysemy and contextual meaning, a particular challenge in fiction, where the same word may carry spatial or non-spatial meaning depending on context. Our classifier, trained on manually annotated prose, implicitly learns contextual meaning rather than relying on surface vocabulary. 

To model patterns of setting, we draw on prior work demonstrating that fine-tuning transformer models on hand-annotated data is effective for capturing complex narratological phenomena, including narrativity, events, and space \cite{antoniak-etal-2024-people, vauth-etal-2021-automated, kababgi2024recognising, soni-etal-2023-grounding}. Unlike dictionary-based methods, transformer classifiers capture contextual and long-range semantic dependencies, allowing for the detection of abstract patterns beyond n-gram overlap \cite{vaswani_attention_2023, wankmuller_introduction_2024}.  

Research on long-form narrative generation remains limited. Existing studies often focus on short passages or synopses \cite{tian_are_2024}.\footnote{Here, we use \textit{long-form} to refer to multi-chapter narratives of several thousand words, as opposed to single passages, synopses, or summaries.}  While LLMs continue to struggle with extended narrative coherence, longer context windows and improved prompting strategies have made extended generation more feasible for recent models (e.g., \citealt{wang-etal-2025-generating, bae_collective_2024}).

\section{Method}

\subsection{Dataset}
We draw a random sample of 1,000 works per language from English and German fiction corpora of approximately 4,300 public-domain texts. The English corpus (1800--1920) was scraped from Project Gutenberg based on the metadata provided by 
\citet{cuthbert_gender_novels_2019}. Genre labels for the English corpus were assigned by us using the subject and topic metadata that Project Gutenberg supplies for each work. The German corpus (1780--1940) was drawn from \citet{rohrbacher2025b}, itself sourced primarily from Projekt Gutenberg-DE with a small fraction coming from German works in the English Project Gutenberg. Genre labels for the German corpus were taken from Projekt Gutenberg-DE, which follows the categories of the German book trade, and are provided with the corpus. Both corpora consist primarily of novels and novellas and cover a broad range of narrative subgenres, including speculative fiction, crime fiction, fairy tales, and young adult literature. 

\subsection{Narrative generation}
\label{sec:generation}
We use prompts that contain the first sentence of each human-written text to generate long-form narratives of four chapters averaging 6,500--9,400 words per model (Appendix~\ref{app:length}). We refer to these model-generated texts as “continuations”. Genre information is provided as an additional conditioning input (e.g., novel, speculative fiction, fairy tale). To analyze variation across model families, we prompted three open-source models and one closed-source model (LlaMA-3.3~70B \cite{grattafiori_llama_2024}, Mistral~3.2~24B\footnote{\url{https://huggingface.co/mistralai/Mistral-Small-3.2-24B-Instruct-2506}}, Gemma~3~27B \cite{gemma3_technical_report_2025}, and GPT~4.1 (gpt-4.1-2025-04-14)\footnote{\url{https://openai.com/index/gpt-4-1/}}).\footnote{For brevity, we henceforth refer to these models as LlaMA~3.3, Mistral~3.2, Gemma~3, and GPT~4.1 respectively.} Model selection was informed by prior evaluations of creative writing performance~\cite{paech2023eqbench}. Additional models were piloted but excluded due to recurrent generation failures, including truncation, repetition, and unintended language switching.\footnote{Excluded models: Mixtral, LlaMA-3-8B, QwQ-32B, Qwen-2.5, Apertus.} We sampled all models with a temperature and top-p of 1 to study model behavior without the influence of sampling parameters, making results comparable across models. For reproducibility, we used a seeded random function with a fixed seed. 

\begin{table}[t]
\centering
\begin{tabular}{lcc}
\toprule
\textbf{Model} & \textbf{Sent. Length} \\ 
\midrule
GPT~4.1 \textit{en}& 16 \\
Gemma~3 \textit{en}& 11 \\

LlaMA 3.3 \textit{en}& 17 \\
Mistral 3.2 \textit{en}& 16 \\
GPT~4.1 \textit{de}& 16 \\
Gemma 3 \textit{de}& 11 \\
Llama 3.3 \textit{de}& 14 \\
Mistral 3.2 \textit{de} & 14 \\
\midrule
Human \textit{en}& 25 \\
Human \textit{de}& 27 \\
\bottomrule
\end{tabular}
\caption{Average sentence length in words for English (\textit{en}) and German (\textit{de}) texts. The matched human excerpts average 10,761 words in English and 7,953 in German. Story lengths for the generated texts are reported in Appendix~\ref{app:length}}
\label{tab:english-data}
\end{table}

Because long-form generation remains challenging for many models, particularly open-source ones, we employ an iterative, chapter-based prompting strategy. Rather than issuing a single prompt, we guide generation incrementally, with each chapter conditioned on the model's prior output. As shown in Table~\ref{tab:prompting}, the model is assigned a system role and given stepwise user instructions for generating the narrative chapters. Prompts were refined through iterative testing to suppress meta-commentary, user-directed dialogue, repetition, and premature endings. The resulting dataset, a corpus of 8,000 AI-generated stories (1,000 per model for each language), is available in our code repository, together with the metadata and opening sentences of the matched human-authored sample.\footnote{Code and data used in this project can be found here: \url{https://github.com/BjoernNieth/worldbuilding-AI}. Dataset details are reported in Table \ref{tab:english-data}.} 

We report four robustness checks. To rule out memorization, we measured 13-gram overlap between generated texts and their source books. To confirm robustness to prompt formulation, we ran three prompt variants for each open-source model and language, which vary phrasing and verbosity (see Appendix~\ref{app:ablation}). Because each chapter is prompted separately, every chapter start may behave like a story opening and drive the patterns we report across narrative time. To test this, we re-bin the generated texts by position within chapter (Section~\ref{sec:analysis-design}). Finally, we include genre as a covariate in the statistical model, since it is supplied to the models during generation and could influence the spatial distributions (Section~\ref{sec:stats}). 

\begin{table*}[ht]
\centering
\small
\renewcommand{\arraystretch}{1.4}
\begin{tabularx}{\textwidth}{>{\raggedright\arraybackslash}p{2.2cm}>{\raggedright\arraybackslash}X}
\toprule
\textbf{Role} & \textbf{Instruction} \\
\midrule
System & You are an award-winning author. Your task is to write a fictional 
narrative. Write a continuous narrative without interruptions or questions. 
Do not break your role as the author. Avoid summarizing or bringing the 
story to an end. Avoid repetitions and do not rewrite text from the previous 
chapter. Each chapter should introduce new elements, deepen existing 
characters, and further develop the plot. \\
\midrule
Step 1 & Write a fictional story in the following genre: \texttt{\{genre\}}. 
Use the following opening: \texttt{\{sentence\}}. Write the first chapter 
with approximately 3,000 tokens. Avoid any AI comments or meta-statements 
that break the role of the author. \\
\midrule
Step 2 \newline \textit{\small (repeated per chapter)} & Continue the narrative 
from where the previous chapter ended, introducing new plot elements and 
deepening existing characters ($\sim$3,000 tokens). Avoid repetition and 
meta-commentary. \textit{Repeated for four chapters.} \\
\bottomrule
\end{tabularx}
\caption{Instructions for model-generated narratives. \texttt{\{genre\}} 
and \texttt{\{sentence\}} are filled with metadata and the opening sentence 
of each source text. Generation was capped at 3,000 tokens per chapter and stopped at the end-of-sequence token, for a total of 10,000--12,000 tokens per story (Appendix~\ref{app:length}).}
\label{tab:prompting}
\end{table*}

\subsection{Setting classification}

We use a fine-tuned BERT classifier that assigns each sentence to one of five categories: action space, perceived space, visual space, descriptive space, and ``no space'' \cite{rohrbacher_opening_2025a}. The five categories capture distinct modes of spatial representation. See Table~\ref{tab:setting-categories} for full definitions and examples. The classifier was originally developed for German-language fiction. For the present study, we replicate it for English by fine-tuning on a manually annotated dataset of English fictional prose following the same annotation scheme, achieving comparable performance (see Appendix~\ref{app:eng-classifier}). Since the classifier was fine-tuned on human-authored fictional prose, we validated its generalization to AI-generated text using a manually annotated sample (see Section~\ref{sec:results-validation}).


\begin{table*}[ht]
\centering
\small
\renewcommand{\arraystretch}{1.4}
\begin{tabularx}{\textwidth}{>{\raggedright\arraybackslash}p{2.2cm}>{\raggedright\arraybackslash}p{5.5cm}>{\raggedright\arraybackslash}X}
\toprule
\textbf{Type} & \textbf{Definition} & \textbf{Example} \\
\midrule
Action space & Space characters move through; objects serve functional roles. & He jumped up, jerked the window-shade, and dragged his chair closer to examine the shoes \citep{thierry_adventure_1918}. \\
\midrule
Perceived space & Space sensorially experienced; contributes to mood and atmosphere. & The terror of loneliness among those overhanging mountains gripped at the boy's throat \citep{garland_ross_grant}. \\
\midrule
Visual space & Space viewed from a static position; presents itself to the character. & He glanced from Tom to the cabin \citep{chapman_tom_fairfield}. \\
\midrule
Descriptive space & Spatial information not tied to character agency. & On either side of the towpath were farms and gardens \citep{hill_corner_house_girls}. \\
\midrule
No space & No spatial relationship present. & By this curious turn of disposition I have gained the reputation of deliberate heartlessness \citep{bronte_wuthering_1847}. \\
\bottomrule
\end{tabularx}
\captionsetup{justification=raggedright, singlelinecheck=false}
\caption{Categories of narrative setting used by the setting classifier (adapted from \citet{rohrbacher2025b}). Labels follow English translations of the original German terms: \textit{Aktionsraum} (action space), \textit{gestimmter Raum} (perceived space), \textit{Anschauungsraum} (visual space).}
\label{tab:setting-categories}
\end{table*}



\subsection{Analysis design}
\label{sec:analysis-design}

To compare how narrative environments are structured across AI-generated and human-authored texts, we compute the proportion of each category across all sentences in a corpus. We apply the classifier to both, matching each literary excerpt to the length of its corresponding generated story.\footnote{\citet{lucy_gender_2021} adopt a comparable design, prompting GPT-3 with opening sentences from novels and comparing the output against length-matched excerpts from the same books.} We analyze setting at two levels of granularity. At the micro level, we focus on the opening of the text. Narratological research identifies openings as the primary site of world-establishment~\cite{gavins_text_nodate, herman_basic_2009}, but they have no fixed textual boundary. We therefore operationalize the opening as the first 15 sentences of each text, an approximation we adopt for the present analysis. At the macro level, we segment texts into ten equal-length sections and track the normalized frequency of each category across these intervals. Here, normalized frequency refers to the proportion of sentences assigned to a given category within each section. This design is motivated by prior work showing that fictional texts follow recognizable patterns in how setting unfolds across narrative time \cite{rohrbacher_opening_2025a, boyd2020narrative}. Although the generated excerpts lack canonical endings, this normalization allows comparison across texts of varying length.

Because narratives were generated chapter by chapter (Section~\ref{sec:generation}), openings recur at every chapter boundary, and the ten-section grid is not aligned to them. We therefore additionally re-bin each generated text by position within chapter, dividing each of the four chapters into quarters (16 bins per story). 



\subsection{Statistical analysis}
\label{sec:stats}
To test whether setting category distributions differ systematically between human-authored and model-generated texts across narrative time, we fit a Generalized Linear Mixed Model (GLMM) for each spatial category using the glmmTMB R package \cite{brooks_glmmtmb_2017}. We modeled each setting category as a function of a five-level author factor (Human, GPT~4.1, Gemma~3, LlaMA~3.3, Mistral~3.2) crossed with narrative section. To account for story-specific variation, the model included a by-story random intercept. Genre was included as a covariate. Due to the small size of most genre types, we code genre as novel vs. other fiction to avoid modeling minor subgenres separately. Each generated story takes the genre label of the source text that seeded it, so the genre distribution is identical across all five author conditions (English: 69.2\% novel; German: 79.6\%). The full specification is space \textasciitilde{} author $\times$ section + genre + (1 | story). As each outcome is a proportion bounded in $[0,1]$, including exact zeros and ones, we used the ordbeta family \cite{kubinec_ordered_2023}. Significance was assessed via Type~III Wald $\chi^2$ tests. Post-hoc contrasts were computed as estimated marginal means using \texttt{emmeans} \cite{lenth_emmeans_2026}.

\section{Results}

\subsection{Classifier validation}
\label{sec:results-validation}

Two annotators (the first author and a research student) manually labeled a stratified sample of 600 sentences drawn from the AI-generated texts, covering all four models and both languages. Inter-annotator agreement was substantial ($\kappa = 0.736$, 79.2\% raw agreement). The classifier achieved an accuracy of 78.0\% and a mean Cohen's $\kappa$ of 0.725, closely approaching the human inter-annotator ceiling. Performance was consistent across languages and models, with the exception of Mistral~3.2 (see Appendix~\ref{app:clf-validation}). 

\subsection{Analysis of setting in story openings}

\begin{figure}[t]
    \centering
    \includegraphics[width=\columnwidth]{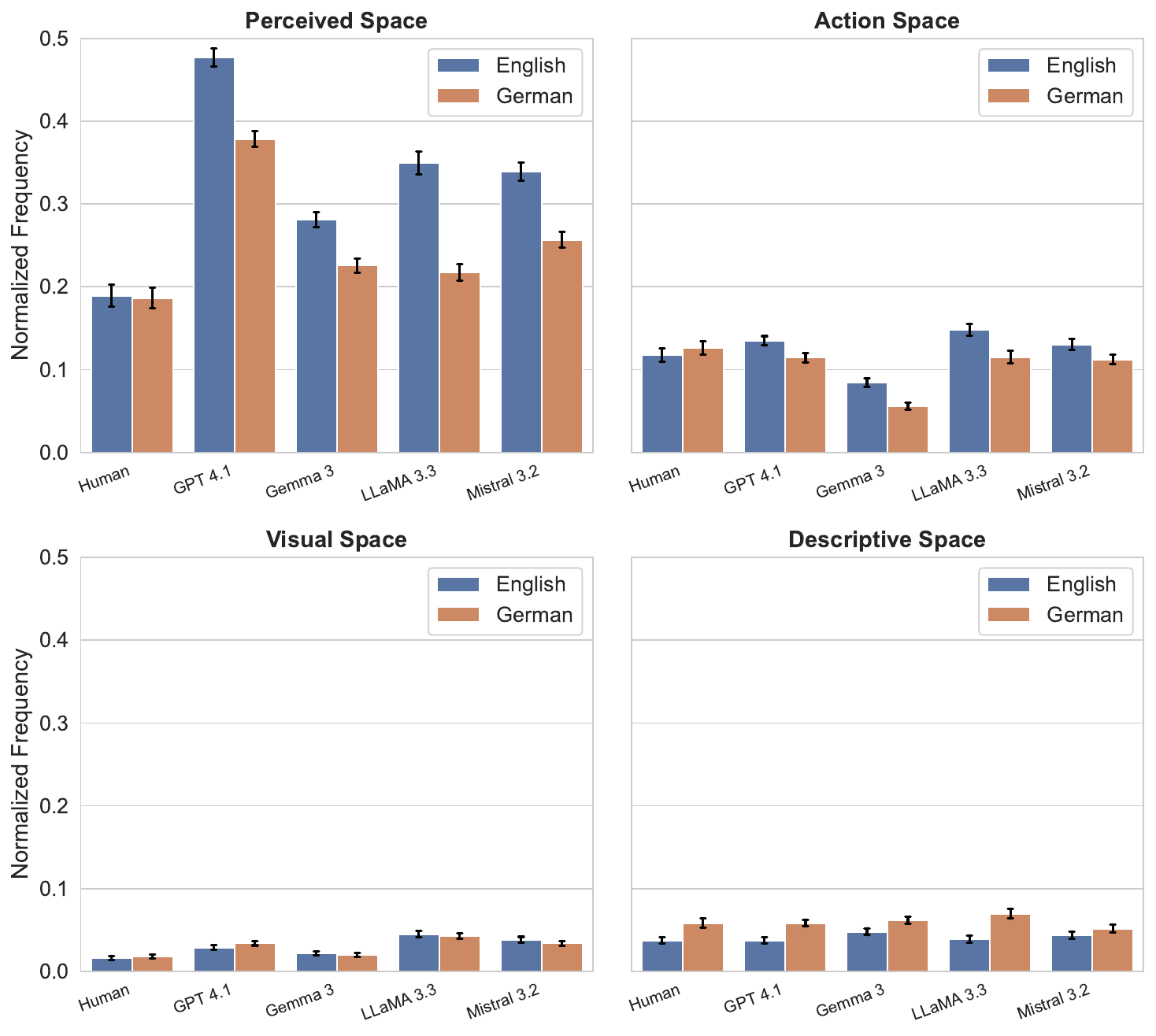}
    \caption{Normalized frequencies of setting categories in the opening sentences (first 15 sentences) of human-authored and AI-generated texts in English and German. 
    Error bars indicate 95\% confidence intervals.}
    \label{fig:opening-frequencies}
\end{figure}

Figure~\ref{fig:opening-frequencies} shows the normalized frequency of each spatial category in the opening 15 sentences, compared across models and 
languages. It reveals a pronounced contrast between human-authored and model-generated openings. Perceived space is the most frequent category in both model-generated texts and human-authored fiction. Whereas human-authored fiction shows normalized frequencies of around 0.19 in both 
languages, all four models exceed this level, with GPT~4.1 reaching 
around 0.47 in English and 0.38 in German, roughly twice the human 
baseline. 
For action space, Gemma~3 differs most strongly 
from the human baseline, while the other three models remain broadly within the same range. Visual space shows a moderate but consistent increase in model-generated text, most 
pronounced for LlaMA~3.3, which reaches roughly twice the human 
baseline in both languages.  Descriptive space is the category that models reproduce most closely, with language-specific differences shown in both human and AI-generated texts. The overall gap between human-authored and generated texts is  more pronounced in English than in German.


\begin{table*}[t]
\small
\begin{minipage}[t]{0.48\textwidth}
  \textbf{Human-authored}\\
  \footnotesize \textit{Frank Norris, McTeague: A Story of San Francisco (1899)}

  \vspace{6pt}

  \spacea{\textbf{It was Sunday, and, according to his custom on that
  day, McTeague took his dinner at two in the afternoon at the
  car conductors' coffee-joint on Polk Street.}}\abadge{}
  \spacep{He had a thick gray soup; heavy, underdone meat, very hot,
  on a cold plate; two kinds of vegetables; and a sort of suet
  pudding, full of strong butter and sugar.}\pbadge{}
  \spacea{On his way back to his office, one block above, he
  stopped at Joe Frenna's saloon and bought a pitcher of steam
  beer.}\abadge{}
  \spacea{It was his habit to leave the pitcher there on his
  way to dinner.}\abadge{}
  \spacea{Once in his office, he took off his coat and shoes,
  unbuttoned his vest, and, having crammed his little stove
  full of coke, lay back in his operating chair at the bay
  window, reading the paper, drinking his beer, and smoking
  his huge porcelain pipe while his food digested; crop-full,
  stupid, and warm.}\abadge{}
  \spacep{By and by, gorged with steam beer, and overcome by
  the heat of the room, the cheap tobacco, and the effects of
  his heavy meal, he dropped off to sleep. [...]}\pbadge{}

\end{minipage}
\hfill
\vrule width 0.4pt
\hfill
\begin{minipage}[t]{0.48\textwidth}
  \textbf{GPT~4.1 generated}\\
  \textit{\footnotesize Continuation from same opening sentence}

  \vspace{6pt}

  \spacea{\textbf{It was Sunday, and, according to his custom on that
  day, McTeague took his dinner at two in the afternoon at the
  car conductors' coffee-joint on Polk Street.}}\abadge{}
  \spacep{The sky outside was a sullen, unpolished grey,
  holding back the barely audible grumble of distant thunder;
  a few straggling shoppers hurried past the big front window,
  bent under coats and unhappy hats, all the world's machinery
  seeming to slow with the prospect of Sunday's late rain.}\pbadge{}
  \spacep{Inside, under an uneven circle of yellow light,
  McTeague sat waiting, his big hands turning and returning a
  battered steel fork, above a square wooden patch of table
  darkened by years of elbows and spilled drinks.}\pbadge{}
  \spacep{The place was narrow, long, heavy with the paraffin
  tang of old oilcloth and the low murmurs of the
  regulars, conductors in soiled blue coats, a few shop
  clerks, a woman with a battered net of oranges, all as
  familiar and faded to McTeague as the blurred
  wallpaper.}\pbadge{}
  \spaced{There were no clocks in the coffee-joint. [...]}\dbadge{}

\end{minipage}

\vspace{8pt}
\hrule
\vspace{5pt}
{\footnotesize
  \textbf{Legend:}\quad
  \colorbox{actionbg}{\textcolor{actionfg}{action space}}\quad
  \colorbox{perceivedbg}{\textcolor{perceivedfg}{perceived space}}\quad
  \colorbox{descbg}{\textcolor{descfg}{descriptive space}}
}
\caption{Comparison of story openings: human-authored passage from Frank Norris' \emph{McTeague: A Story of San Francisco} \citep{norris_mcteague_1899} versus GPT~4.1 continuation of the same opening sentence. Colours indicate the narrative space type assigned by the setting classifier.}
\label{tab:story-comparison}
\end{table*}

The example in 
Table~\ref{tab:story-comparison} illustrates how these distributional 
differences operate at the sentence level. After the shared opening 
sentence, the two passages immediately diverge. Norris constructs
space primarily through movement and touch. The character McTeague traverses a habitual route from restaurant to saloon to office, and the setting emerges through that trajectory. Objects serve functional roles, the 
pitcher left at the saloon to be collected on the way back, the 
stove crammed with coke. Perceived space, where it appears, is bound to the body and conveys temperature and smoke only through their effect on the character. GPT~4.1, by contrast, arrests movement from the second sentence onward. Three consecutive sentences of perceived space follow. Objects become prominent as sensory props rather than instruments for purposeful action, the ``turning and returning'' of the fork, the table textured by ``years'' of use. The environment is 
rendered quasi-anthropomorphic, its grey sky ``sullen'' and ``the world's machinery seeming to slow.'' Unlike in Norris' passage, the detail of the external world is emphasized. The passage illustrates a storyworld pervaded by atmosphere, where the environment presses in on the character. What the classifier labels as perceived space is, at the level of narrative experience, a world felt before it is actively inhabited. 

\subsection{Temporal distribution of setting across narrative time}
\label{sec:temporal}

\subsubsection{Overall differences in proportion}

\begin{figure*}[ht]
    \centering
    \includegraphics[width=\textwidth]{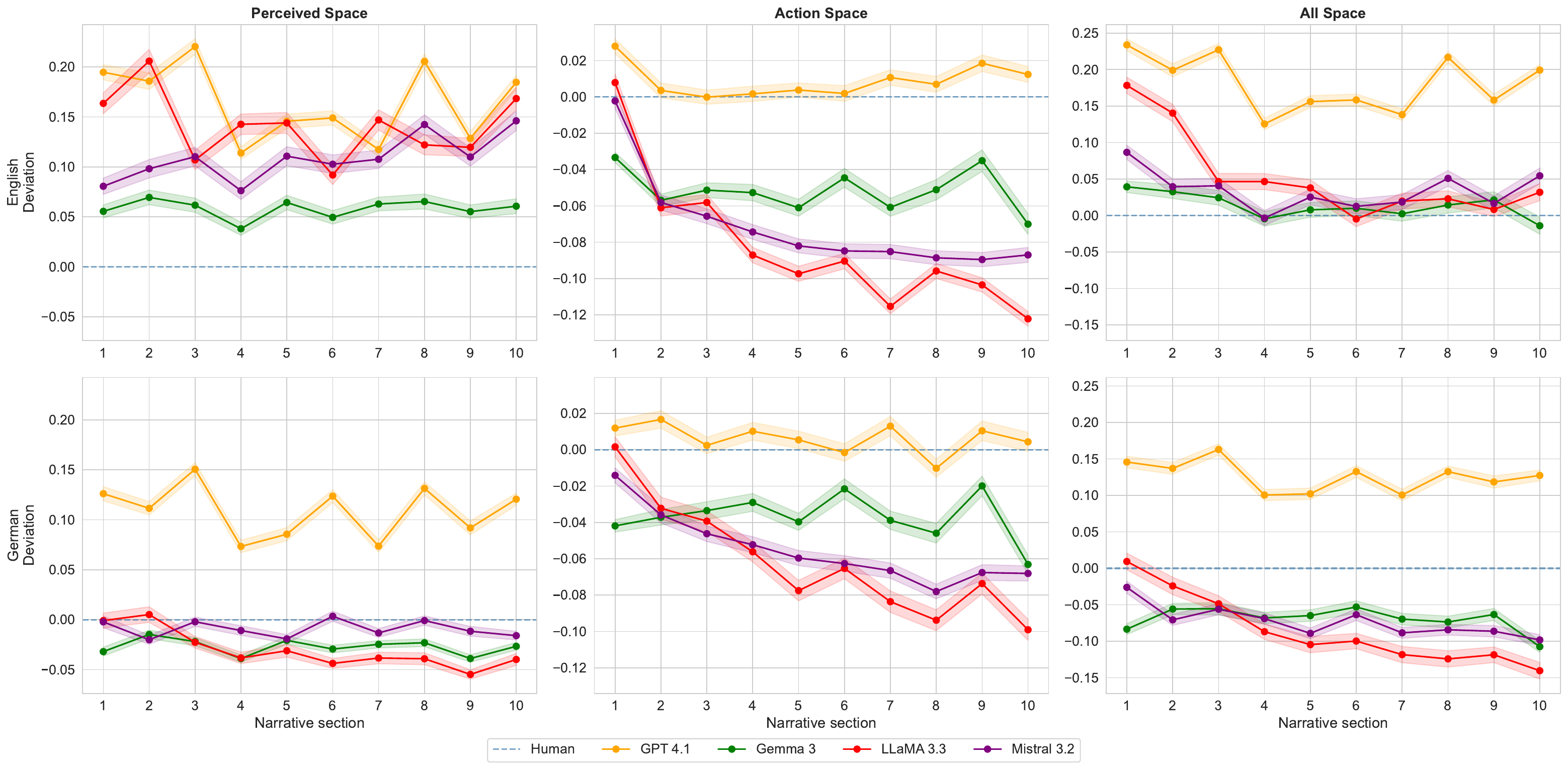}
    \caption{Deviation of model predictions from the human baseline for perceived and action space across narrative sections, comparing English and German corpora.}
    \label{fig:dev-bilingual}
\end{figure*}

In human-authored fiction, action space is the most frequently produced spatial category. LLMs deviate systematically from the distributional profile of human-authored fiction, most consistently in their overuse of perceived space. Figure~\ref{fig:dev-bilingual} shows the deviation of each model from the human baseline across ten narrative sections for English and German. The statistical model confirms significant model differences in the first section and a significant model$\times$section interaction, indicating that models diverge from human authors in ways that vary across narrative time (English: model main effect $\chi^2(4) = 2509.37$, $p < .001$; model$\times$section interaction $\chi^2(36) = 2013.48$, $p < .001$). Post-hoc contrasts confirm that all four LLMs produce significantly more perceived space than human authors across every narrative section (all $p < .001$ after Holm correction). GPT~4.1 shows the largest deviations on average (${\sim}0.14$--$0.25$ above baseline) and LlaMA~3.3 the second largest (${\sim}0.10$--$0.23$), while Gemma~3 remains closest to the human baseline (${\sim}0.06$--$0.10$). For action space, GPT~4.1 is the only model that remains close to or above the human baseline. Gemma~3 and Mistral~3.2 fall below it from the opening section onward, while LlaMA~3.3 diverges from section 2 onward. These differences produce a significant model effect (English: $\chi^2(4) = 456.78$, $p < .001$). LlaMA~3.3 shows the largest deficit overall, based on post-hoc contrasts (estimates $-0.04$\,--\,$-0.07$
across sections ~2--10, all $p < .001$). This pattern is reflected in the all-space aggregate: GPT~4.1 produces a consistently higher proportion of spatially marked sentences than human authors. 

\subsubsection{Differences in trend}
\label{sec:trend}

\begin{figure*}[ht]
    \centering
    \includegraphics[width=\textwidth]{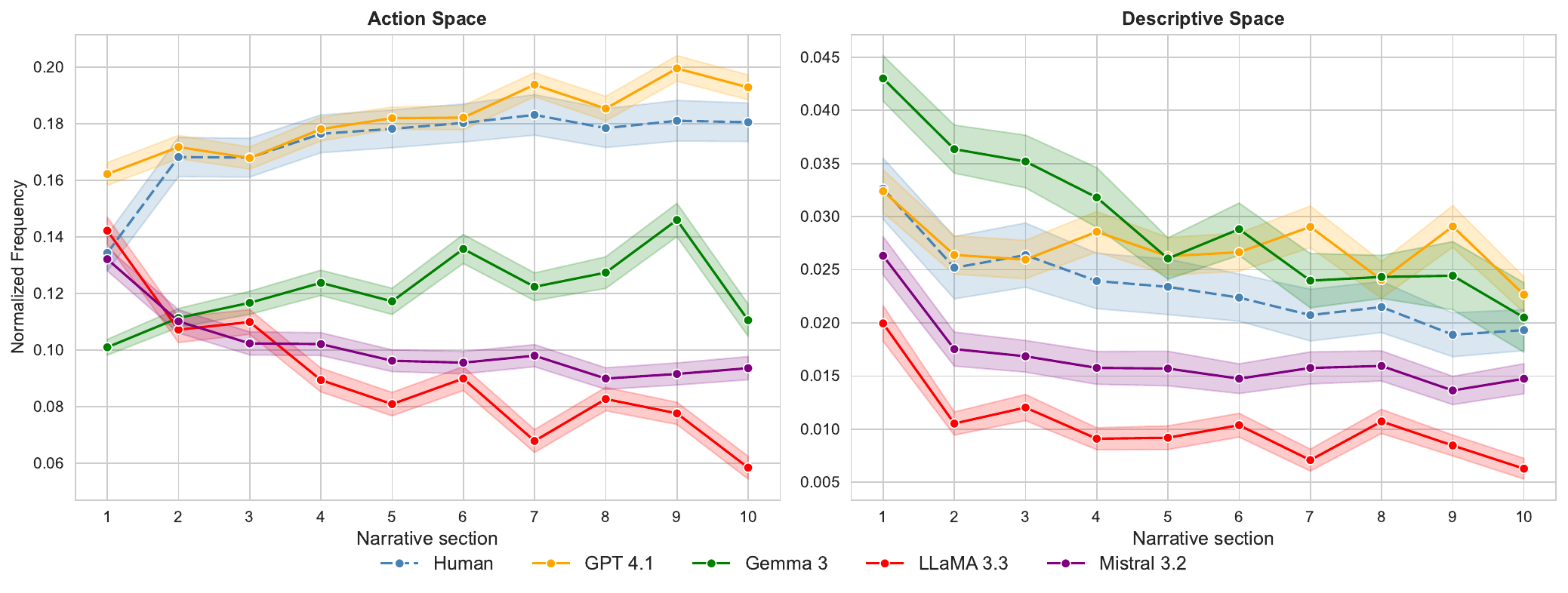}
    \caption{Normalized frequency of \textit{action space} and \textit{descriptive space}  across narrative 
    sections for human-authored and 
    AI-generated texts in English. Shaded bands indicate $\pm$1 standard 
    error.}
    \label{fig:arc-spotlight}
\end{figure*}

Figure~\ref{fig:arc-spotlight} shows action and descriptive space frequency across ten narrative sections. For action space, GPT~4.1 tracks closely with the human baseline and shares its slight upward trend. LlaMA~3.3 declines continuously, while Gemma~3 and Mistral~3.2 stabilize below the baseline after an initial drop. For descriptive space, all models broadly follow the human baseline's declining trend, with GPT~4.1 and Gemma~3 remaining somewhat elevated, and Mistral~3.2 and LlaMA~3.3 falling below it. For perceived space (Figure~\ref{fig:five-boxes} in Appendix~\ref{app:temporal-dist}), the human baseline starts moderately and remains largely flat across narrative sections, whereas all models start higher and show no comparable stabilization. GPT~4.1 and LlaMA~3.3 in particular fluctuate considerably throughout. Part of this variance is attributable to chapter boundaries in the generation procedure (Section~\ref{sec:ablations}, Appendix~\ref{app:chapters}). Beyond the boundary effect, the models distribute atmospheric density less evenly across a narrative than human authors do. 

\subsubsection{Comparison English vs. German}

The cross-linguistic comparison (Figure~\ref{fig:dev-bilingual}) shows a reversal for perceived space. While all models exceed the English human baseline, most fall at or below the German baseline (GPT~4.1 excepted, which remains elevated in both languages). Post-hoc contrasts show that Gemma~3 and Mistral~3.2 are indistinguishable from the German human baseline in several later sections,  while LlaMA~3.3 falls consistently below it from section~3 onward ($\chi^2(36) = 1860.15$, $p < .001$). For action space, the pattern of deviation replicates across languages but is more pronounced in German ($\chi^2(36) = 1449.79$, $p < .001$). LlaMA's declining trajectory and Gemma's and Mistral's stable deficit are visible in both corpora, with the gap widening in German. The all-space aggregate mirrors this. GPT~4.1 exceeds the human baseline in both languages, whereas the remaining 
models, which approach human levels in English, fall clearly below it in German.

\subsection{Ablation studies}
\label{sec:ablations}

\paragraph{Memorization.} Across both languages we found no instances of verbatim reconstruction, with the single exception of Gemma~3 reproducing the well-known opening sentence of Dickens' \textit{A Tale of Two Cities}.
\paragraph{Prompt sensitivity.} As shown in Figures~\ref{fig:ablation-sections-en} and~\ref{fig:ablation-sections-ger}, prompt formulation influences the absolute frequency of spatial categories to some degree, but overall trajectories across narrative sections remain consistent across all conditions, confirming that the main findings are not an artifact of the specific prompt used.
\paragraph{Chapter-boundary position.} Perceived space is elevated at the start of every chapter relative to the second quarter of the chapter, across all eight model-language conditions, and rises again in the final quarter in English (except Gemma 3) and, in German, for GPT~4.1. The effect is small relative to the difference we report. The largest within-chapter range is 0.09, whereas the English models exceed the human baseline by 0.11 to 0.21 overall, and even at their within-chapter minimum all four produce two to three times as much perceived space as human authors. The overproduction is moreover already present in chapter 1, which is generated from a single prompt before any continuation instruction. Part of the temporal variance described in Appendix~\ref{app:temporal-dist} is therefore attributable to the chapter-based prompting strategy, but the overproduction itself is not. 
\paragraph{Genre.} Genre is significant for perceived and action space in English ($\chi^2(1) = 18.63$ and $23.14$, both $p < .001$) and for all categories except descriptive space in German. The author main effect and the author\,$\times$\, section interaction remain highly significant with genre included (Appendix~\ref{app:glmm}). 

\section{Discussion}

In human-authored fiction, action space in story openings anchors readers in sequences of habitual, embodied action, generating narrative momentum through characters' movement and routine interaction with space. 
LLM-generated openings, instead, privilege perceived space, foregrounding \textit{Stimmung} (i.e., mood or atmosphere) over action, resulting in a diffuse “background feeling” \cite{colombetti_feeling_2013} untethered from what characters actually do.\footnote{The original German term is \textit{gestimmter Raum}, i.e. space as it reflects atmosphere or mood, which more directly captures the connotation of space as permeated by feeling, a nuance that the English “perceived space” only partially conveys.}
This skew contributes to an empirically recognizable generative AI style---rich in affect and ambience, but less grounded in embodied action and spatial concreteness. This pattern aligns with prior work documenting LLMs' preference for sensorial and affect-laden language \cite{hicke_zero_2025, lee_language_2024}, and extends it by showing how these preferences manifest at the level of narrative setting and shape storyworld construction. 

A plausible explanation is that post-training alignment methods might amplify this tendency. If human raters reward emotionally resonant, atmospheric prose, generation may be biased systematically toward perceived space. Importantly, the pattern we observe does not appear to be driven by historical periodicity in the human baseline. Spatial category distributions in the human-authored corpora show no systematic directional trend across the main period of the corpus (see Appendix~\ref{app:temporal}). This suggests the observed differences reflect a systematic contrast between human and LLM writing rather than a temporal drift in narrative space composition.

A second explanation concerns the generation procedure itself. Because each chapter is prompted afresh, every chapter start is treated as an opening, and perceived space rises at chapter boundaries relative to chapter middles (Appendix~\ref{app:chapters}). This accounts for part of the fluctuation across narrative sections, but not for the overall level, which is already elevated in chapter~1. Chapter-level prompting is a necessity given current long-form performance, and a different strategy would produce a different boundary profile without removing the underlying skew.

The cross-linguistic asymmetry suggests that these tendencies are not uniformly stable across languages. Open-source models align more closely with GPT~4.1 in English than in German. In the latter, their perceived space distributions converge toward or fall below the human baseline, and their action space deficit widens. The perceived and action spaces of GPT~4.1, by contrast, remain elevated in both languages. In most cases, the three open-source models show comparatively little variation relative to each other despite differing in architecture and size. This suggests that the observed spatial tendencies reflect shared properties of instruction-tuned LLMs more broadly, such as preference optimization or training data composition, rather than model-specific factors. 

Included as a covariate, genre has a significant effect for several categories in both languages (Appendix~\ref{app:glmm}). But because genre is matched across author conditions, this does not affect the human-model comparison. It does, however, indicate that spatial composition varies with genre, and the framework could be applied to genre-stratified corpora to examine this directly. 
 

The fact that spatial category distributions alone are sufficient to identify the generating model well above chance (see Appendix~\ref{app:classifier}) indicates that setting constitutes a reliable stylistic marker of AI-generated fiction. We identify two directions that could build on this. The first is space-conditioned generation, in which models would be explicitly constrained to produce texts belonging to certain narrative spaces during prompting, which could mitigate the differences we report. The second is empirical validation with readers. Our analysis documents a difference in spatial composition, which is manifest in the texts themselves, but does not establish what that difference means for the reading experience. Existing work compares human-authored and AI-generated stories as wholes and finds that readers rate the AI-generated ones as more absorbing \citep{sears_bot_2026}, but these judgements are not linked to any measured property of the texts. A reader study could vary spatial composition directly and test whether it corresponds to differences in immersion or perceived quality. 


\section{Conclusion}

We introduced a narratology-informed framework for analyzing worldbuilding in AI-generated fiction, comprising five spatial categories grounded in narrative theory, a fine-tuned BERT classifier for English (extending an existing German classifier), and a corpus-scale application across four LLMs in both languages. The results reveal consistent differences between LLM-generated and human-authored text in how narrative space is constructed. LLMs systematically overrepresent perceived space while producing less action space than human authors, a pattern that holds across models and across narrative sections, though with model-specific and cross-linguistic variation in magnitude. GPT~4.1 shows the largest and most consistent overproduction of perceived space across both languages, while remaining close to the human baseline for action and descriptive space. These findings suggest that LLMs do not reproduce the spatial distributions of human fiction, but construct storyworlds in a way that differs systematically from literary norms. The spatial metrics that we propose in this paper, derived from narratological theory and grounded in literary scholarship, offer a path toward richer evaluation of AI-generated narratives.

\section*{Limitations}

Long-form generation remains challenging for LLMs, and the patterns we report, particularly regarding narrative time, may therefore reflect tendencies rather than fully stable narrative strategies. Because we prompted the models to produce “chapters”, part of the variance in spatial distributions across narrative time is attributable to chapter boundaries, where perceived space is elevated (Section~\ref{sec:ablations}, Appendix~\ref{app:chapters}). The classifier's comparatively lower performance on Mistral~3.2 outputs, most pronounced in German, means that the results for this model should be interpreted with some caution. Additionally, our analysis focuses on English and German texts from public-domain Project Gutenberg (English: 1800--1920; German: 1780--1940), which may limit the generalizability of our findings to contemporary fiction and to languages beyond English and German. Extending the analysis to contemporary fiction remains an open direction, and would show whether the patterns we report hold for present-day fiction. Since we use only the first sentence of each human-authored text as a prompt and provide no information on year of publication, we cannot expect LLMs to reproduce the stylistic conventions of this historical period.\footnote{Recent research has shown that LLMs struggle to reproduce the “period style” \cite{underwood_can_2025} of fictional prose when specifically prompted with examples from that period.} Our coarse novel vs.\ other coding cannot capture genre-specific worldbuilding strategies (e.g., “speculative fiction” vs.\ “fairy tales”), and subgenre-stratified sample sizes are too small relative to the dominant novel and novella categories to test whether the divergence we report varies in magnitude across genres. A genre-balanced corpus would be needed to answer this, which we leave to future work. Despite these limitations, our results provide valuable insight into systematic differences between AI-generated and human-authored fiction.

\section*{Acknowledgments}
The authors gratefully acknowledge the scientific support and HPC resources provided by the Erlangen National High Performance Computing Center (NHR@FAU) of the Friedrich-Alexander-Universität Erlangen-Nürnberg (FAU). The hardware is partially funded by the German Research Foundation (DFG). The authors further acknowledge support from the Alexander von Humboldt Foundation as part of the Alexander von Humboldt Professorship endowed by the German Federal Ministry of Research, Technology and Space (BMFTR). The authors thank Jan-Oliver Reincke for help with the annotation. 

\bibliography{custom}

\begin{thebibliography}{52}
\providecommand{\natexlab}[1]{#1}

\bibitem[{Ahuja et~al.(2025)Ahuja, Sclar, and Tsvetkov}]{ahuja2025finding}
Kabir Ahuja, Melanie Sclar, and Yulia Tsvetkov. 2025.
\newblock Finding flawed fictions: Evaluating complex reasoning in language models via plot hole detection.
\newblock \emph{Preprint, arXiv:2504.11900}.

\bibitem[{Antoniak et~al.(2024)Antoniak, Mire, Sap, Ash, and Piper}]{antoniak-etal-2024-people}
Maria Antoniak, Joel Mire, Maarten Sap, Elliott Ash, and Andrew Piper. 2024.
\newblock \href {https://doi.org/10.18653/v1/2024.acl-long.383} {Where do people tell stories online? story detection across online communities}.
\newblock In \emph{Proceedings of the 62nd Annual Meeting of the Association for Computational Linguistics (Volume 1: Long Papers)}, pages 7104--7130, Bangkok, Thailand. Association for Computational Linguistics.

\bibitem[{Bae and Kim(2024)}]{bae_collective_2024}
Minwook Bae and Hyounghun Kim. 2024.
\newblock \href {https://doi.org/0.18653/v1/2024.emnlp-main.1046} {Collective critics for creative story generation}.
\newblock In \emph{Proceedings of the 2024 Conference on Empirical Methods in Natural Language Processing}, pages 18784--18819, Miami, Florida, USA. Association for Computational Linguistics.

\bibitem[{Boyd et~al.(2020)Boyd, Blackburn, and Pennebaker}]{boyd2020narrative}
Ryan~L Boyd, Kate~G Blackburn, and James~W Pennebaker. 2020.
\newblock The narrative arc: Revealing core narrative structures through text analysis.
\newblock \emph{Science advances}, 6(32):eaba2196.

\bibitem[{Bront{\"e}(1847)}]{bronte_wuthering_1847}
Emily Bront{\"e}. 1847.
\newblock \href {https://www.gutenberg.org/ebooks/768} {\emph{Wuthering Heights}}.
\newblock Project Gutenberg.

\bibitem[{Brooks et~al.(2017)Brooks, Kristensen, van Benthem, Magnusson, Berg, Nielsen, Skaug, M{\ae}chler, and Bolker}]{brooks_glmmtmb_2017}
Mollie~E. Brooks, Kasper Kristensen, Koen~J. van Benthem, Arni Magnusson, Casper~W. Berg, Anders Nielsen, Hans~J. Skaug, Martin M{\ae}chler, and Benjamin~M. Bolker. 2017.
\newblock \href {https://doi.org/10.32614/RJ-2017-066} {{glmmTMB} balances speed and flexibility among packages for zero-inflated generalized linear mixed modeling}.
\newblock \emph{The R Journal}, 9(2):378--400.

\bibitem[{Chakrabarty et~al.(2025)Chakrabarty, Ginsburg, and Dhillon}]{chakrabarty_readers_2025}
Tuhin Chakrabarty, Jane~C. Ginsburg, and Paramveer Dhillon. 2025.
\newblock \href {https://doi.org/10.2139/ssrn.5606570} {Readers {Prefer} {Outputs} of {AI} {Trained} on {Copyrighted} {Books} over {Expert} {Human} {Writers}}.

\bibitem[{Chakrabarty et~al.(2024)Chakrabarty, Laban, Agarwal, Muresan, and Wu}]{chakrabarty_art_2024}
Tuhin Chakrabarty, Philippe Laban, Divyansh Agarwal, Smaranda Muresan, and Chien-Sheng Wu. 2024.
\newblock \href {https://doi.org/10.48550/arXiv.2309.14556} {Art or {Artifice}? {Large} {Language} {Models} and the {False} {Promise} of {Creativity}}.
\newblock \emph{arXiv preprint}.
\newblock ArXiv:2309.14556 [cs].

\bibitem[{Chapman(1910)}]{chapman_tom_fairfield}
Allen Chapman. 1910.
\newblock \href {https://www.gutenberg.org/ebooks/44457} {\emph{Tom Fairfield's Hunting Trip, or Lost in the Wilderness}}.
\newblock Project Gutenberg.

\bibitem[{Chen and Ding(2023)}]{chen_probing_2023}
Honghua Chen and Nai Ding. 2023.
\newblock \href {https://doi.org/10.18653/v1/2023.findings-emnlp.858} {Probing the “{Creativity}” of {Large} {Language} {Models}: {Can} models produce divergent semantic association?}
\newblock In \emph{Findings of the {Association} for {Computational} {Linguistics}: {EMNLP} 2023}, pages 12881--12888, Singapore. Association for Computational Linguistics.

\bibitem[{Cicchetti and Feinstein(1990)}]{cicchetti1990}
Domenic~V. Cicchetti and Alvan~R. Feinstein. 1990.
\newblock \href {https://doi.org/10.1016/0895-4356(90)90159-M} {High agreement but low kappa: {II}. resolving the paradoxes}.
\newblock \emph{Journal of Clinical Epidemiology}, 43(6):551--558.

\bibitem[{Colombetti(2013)}]{colombetti_feeling_2013}
Giovanna Colombetti. 2013.
\newblock \href {http://site.ebrary.com/id/11204352} {\emph{The Feeling Body: Affective Science Meets the Enactive Mind}}.
\newblock MIT Press, Cambridge, MA.

\bibitem[{Cuthbert et~al.(2019)Cuthbert, Tagliaferri, Risi et~al.}]{cuthbert_gender_novels_2019}
Michael Scott~Asato Cuthbert, Lisa Tagliaferri, Stephan Risi, and 1 others. 2019.
\newblock \href {http://gendernovels.digitalhumanitiesmit.org} {Computational reading of gender in novels, 1770--1922}.
\newblock MIT Digital Humanities Lab.

\bibitem[{Devlin et~al.(2019)Devlin, Chang, Lee, and Toutanova}]{devlin_bert_2019}
Jacob Devlin, Ming-Wei Chang, Kenton Lee, and Kristina Toutanova. 2019.
\newblock \href {https://doi.org/10.18653/v1/N19-1423} {{BERT}: {Pre}-training of {Deep} {Bidirectional} {Transformers} for {Language} {Understanding}}.
\newblock In \emph{Proceedings of the 2019 {Conference} of the {North} {American} {Chapter} of the {Association} for {Computational} {Linguistics}: {Human} {Language} {Technologies}, {Volume} 1 ({Long} and {Short} {Papers})}, pages 4171--4186, Minneapolis, Minnesota. Association for Computational Linguistics.

\bibitem[{Fatemi et~al.(2024)Fatemi, Kazemi, Tsitsulin, Malkan, Yim, Palowitch, Seo, Halcrow, and Perozzi}]{fatemi2024test}
Bahare Fatemi, Mehran Kazemi, Anton Tsitsulin, Karishma Malkan, Jinyeong Yim, John Palowitch, Sungyong Seo, Jonathan Halcrow, and Bryan Perozzi. 2024.
\newblock Test of time: A benchmark for evaluating llms on temporal reasoning.
\newblock \emph{arXiv preprint arXiv:2406.09170}.

\bibitem[{Garland(1915)}]{garland_ross_grant}
John Garland. 1915.
\newblock \href {https://www.gutenberg.org/ebooks/34296} {\emph{Ross Grant Tenderfoot}}.
\newblock Project Gutenberg.

\bibitem[{Gavins(2007)}]{gavins_text_nodate}
Joanna Gavins. 2007.
\newblock \emph{Text world theory : an introduction}.
\newblock Edinburgh University Press, Edinburgh.
\newblock OCLC: 1162422251.

\bibitem[{{Gemma Team}(2025)}]{gemma3_technical_report_2025}
{Gemma Team}. 2025.
\newblock \href {https://arxiv.org/abs/2503.19786} {Gemma 3 technical report}.
\newblock ArXiv:2503.19786 [cs.CL].

\bibitem[{Grattafiori et~al.(2024)Grattafiori, Dubey, Jauhri, Pandey, Kadian, Al-Dahle, Letman, Mathur, Schelten, Vaughan, Yang, Fan, Goyal, Hartshorn, Yang, Mitra, Sravankumar, Korenev, Hinsvark, Rao, Zhang, Rodriguez, Gregerson, Spataru, Roziere, Biron, Tang, Chern, Caucheteux, Nayak, Bi, Marra, McConnell, Keller, Touret, Wu, Wong, Ferrer, Nikolaidis, Allonsius, Song, Pintz, Livshits, Wyatt, Esiobu, Choudhary, Mahajan, Garcia-Olano, Perino, Hupkes, Lakomkin, AlBadawy, Lobanova, Dinan, Smith, Radenovic, Guzmán, Zhang, Synnaeve, Lee, Anderson, Thattai, Nail, Mialon, Pang, Cucurell, Nguyen, Korevaar, Xu, Touvron, Zarov, Ibarra, Kloumann, Misra, Evtimov, Zhang, Copet, Lee, Geffert, Vranes, Park, Mahadeokar, Shah, Linde, Billock, Hong, Lee, Fu, Chi, Huang, Liu, Wang, Yu, Bitton, Spisak, Park, Rocca, Johnstun, Saxe, Jia, Alwala, Prasad, Upasani, Plawiak, Li, Heafield, Stone, El-Arini, Iyer, Malik, Chiu, Bhalla, Lakhotia, Rantala-Yeary, Maaten, Chen, Tan, Jenkins, Martin, Madaan, Malo, Blecher, Landzaat,
  Oliveira, Muzzi, Pasupuleti, Singh, Paluri, Kardas, Tsimpoukelli, Oldham, Rita, Pavlova, Kambadur, Lewis, Si, Singh, Hassan, Goyal, Torabi, Bashlykov, Bogoychev, Chatterji, Zhang, Duchenne, Çelebi, Alrassy, Zhang, Li, Vasic, Weng, Bhargava, Dubal, Krishnan, Koura, Xu, He, Dong, Srinivasan, Ganapathy, Calderer, Cabral, Stojnic, Raileanu, Maheswari, Girdhar, Patel, Sauvestre, Polidoro, Sumbaly, Taylor, Silva, Hou, Wang, Hosseini, Chennabasappa, Singh, Bell, Kim, Edunov, Nie, Narang, Raparthy, Shen, Wan, Bhosale, Zhang, Vandenhende, Batra, Whitman, Sootla, Collot, Gururangan, Borodinsky, Herman, Fowler, Sheasha, Georgiou, Scialom, Speckbacher, Mihaylov, Xiao, Karn, Goswami, Gupta, Ramanathan, Kerkez, Gonguet, Do, Vogeti, Albiero, Petrovic, Chu, Xiong, Fu, Meers, Martinet, Wang, Wang, Tan, Xia, Xie, Jia, Wang, Goldschlag, Gaur, Babaei, Wen, Song, Zhang, Li, Mao, Coudert, Yan, Chen, Papakipos, Singh, Srivastava, Jain, Kelsey, Shajnfeld, Gangidi, Victoria, Goldstand, Menon, Sharma, Boesenberg, Baevski,
  Feinstein, Kallet, Sangani, Teo, Yunus, Lupu, Alvarado, Caples, Gu, Ho, Poulton, Ryan, Ramchandani, Dong, Franco, Goyal, Saraf, Chowdhury, Gabriel, Bharambe, Eisenman, Yazdan, James, Maurer, Leonhardi, Huang, Loyd, Paola, Paranjape, Liu, Wu, Ni, Hancock, Wasti, Spence, Stojkovic, Gamido, Montalvo, Parker, Burton, Mejia, Liu, Wang, Kim, Zhou, Hu, Chu, Cai, Tindal, Feichtenhofer, Gao, Civin, Beaty, Kreymer, Li, Adkins, Xu, Testuggine, David, Parikh, Liskovich, Foss, Wang, Le, Holland, Dowling, Jamil, Montgomery, Presani, Hahn, Wood, Le, Brinkman, Arcaute, Dunbar, Smothers, Sun, Kreuk, Tian, Kokkinos, Ozgenel, Caggioni, Kanayet, Seide, Florez, Schwarz, Badeer, Swee, Halpern, Herman, Sizov, Guangyi, Zhang, Lakshminarayanan, Inan, Shojanazeri, Zou, Wang, Zha, Habeeb, Rudolph, Suk, Aspegren, Goldman, Zhan, Damlaj, Molybog, Tufanov, Leontiadis, Veliche, Gat, Weissman, Geboski, Kohli, Lam, Asher, Gaya, Marcus, Tang, Chan, Zhen, Reizenstein, Teboul, Zhong, Jin, Yang, Cummings, Carvill, Shepard, McPhie, Torres,
  Ginsburg, Wang, Wu, U, Saxena, Khandelwal, Zand, Matosich, Veeraraghavan, Michelena, Li, Jagadeesh, Huang, Chawla, Huang, Chen, Garg, A, Silva, Bell, Zhang, Guo, Yu, Moshkovich, Wehrstedt, Khabsa, Avalani, Bhatt, Mankus, Hasson, Lennie, Reso, Groshev, Naumov, Lathi, Keneally, Liu, Seltzer, Valko, Restrepo, Patel, Vyatskov, Samvelyan, Clark, Macey, Wang, Hermoso, Metanat, Rastegari, Bansal, Santhanam, Parks, White, Bawa, Singhal, Egebo, Usunier, Mehta, Laptev, Dong, Cheng, Chernoguz, Hart, Salpekar, Kalinli, Kent, Parekh, Saab, Balaji, Rittner, Bontrager, Roux, Dollar, Zvyagina, Ratanchandani, Yuvraj, Liang, Alao, Rodriguez, Ayub, Murthy, Nayani, Mitra, Parthasarathy, Li, Hogan, Battey, Wang, Howes, Rinott, Mehta, Siby, Bondu, Datta, Chugh, Hunt, Dhillon, Sidorov, Pan, Mahajan, Verma, Yamamoto, Ramaswamy, Lindsay, Lindsay, Feng, Lin, Zha, Patil, Shankar, Zhang, Zhang, Wang, Agarwal, Sajuyigbe, Chintala, Max, Chen, Kehoe, Satterfield, Govindaprasad, Gupta, Deng, Cho, Virk, Subramanian, Choudhury, Goldman,
  Remez, Glaser, Best, Koehler, Robinson, Li, Zhang, Matthews, Chou, Shaked, Vontimitta, Ajayi, Montanez, Mohan, Kumar, Mangla, Ionescu, Poenaru, Mihailescu, Ivanov, Li, Wang, Jiang, Bouaziz, Constable, Tang, Wu, Wang, Wu, Gao, Kleinman, Chen, Hu, Jia, Qi, Li, Zhang, Zhang, Adi, Nam, Yu, Wang, Zhao, Hao, Qian, Li, He, Rait, DeVito, Rosnbrick, Wen, Yang, Zhao, and Ma}]{grattafiori_llama_2024}
Aaron Grattafiori, Abhimanyu Dubey, Abhinav Jauhri, Abhinav Pandey, Abhishek Kadian, Ahmad Al-Dahle, Aiesha Letman, Akhil Mathur, Alan Schelten, Alex Vaughan, Amy Yang, Angela Fan, Anirudh Goyal, Anthony Hartshorn, Aobo Yang, Archi Mitra, Archie Sravankumar, Artem Korenev, Arthur Hinsvark, and 542 others. 2024.
\newblock \href {https://doi.org/10.48550/arXiv.2407.21783} {The {Llama} 3 {Herd} of {Models}}.
\newblock \emph{arXiv preprint}.
\newblock ArXiv:2407.21783 [cs].

\bibitem[{Herman(2002)}]{herman_story_2002}
David Herman. 2002.
\newblock \emph{Story Logic: Problems and Possibilities of Narrative}.
\newblock University of Nebraska Press, Lincoln and London.

\bibitem[{Herman(2009)}]{herman_basic_2009}
David Herman. 2009.
\newblock \emph{Basic elements of narrative}.
\newblock Wiley-Blackwell, Chichester, U.K. ; Malden, MA.
\newblock OCLC: 229467488.

\bibitem[{Hicke et~al.(2025)Hicke, Hamilton, and Mimno}]{hicke_zero_2025}
Rebecca M.~M. Hicke, Sil Hamilton, and David Mimno. 2025.
\newblock \href {https://doi.org/10.48550/arXiv.2504.06393} {The {Zero} {Body} {Problem}: {Probing} {LLM} {Use} of {Sensory} {Language}}.
\newblock \emph{arXiv preprint}.
\newblock ArXiv:2504.06393 [cs].

\bibitem[{Hill(1916)}]{hill_corner_house_girls}
Grace~Brooks Hill. 1916.
\newblock \href {https://www.gutenberg.org/ebooks/38609} {\emph{The Corner House Girls on a Houseboat}}.
\newblock Project Gutenberg.

\bibitem[{Hoffmann(1978)}]{hoffmann1980raum}
Gerhard Hoffmann. 1978.
\newblock \emph{Gerhard Hoffmann: Raum, Situation, erz{\"a}hlte Wirklichkeit. Poetologische und historische Studien zum englischen und amerikanischen Roman.}
\newblock J. B. Metzler.

\bibitem[{Hripcsak and Rothschild(2005)}]{hripcsak2005}
George Hripcsak and Adam~S. Rothschild. 2005.
\newblock \href {https://doi.org/10.1197/jamia.M1733} {Agreement, the {F}-measure, and reliability in information retrieval}.
\newblock \emph{Journal of the American Medical Informatics Association}, 12(3):296--298.

\bibitem[{Ishikawa and Yoshino(2025)}]{ishikawa_ai_2025}
Shin-nosuke Ishikawa and Atsushi Yoshino. 2025.
\newblock \href {https://doi.org/10.18653/v1/2025.nlp4dh-1.51} {{AI} with {Emotions}: {Exploring} {Emotional} {Expressions} in {Large} {Language} {Models}}.
\newblock In \emph{Proceedings of the 5th {International} {Conference} on {Natural} {Language} {Processing} for {Digital} {Humanities}}, pages 614--627, Albuquerque, USA. Association for Computational Linguistics.

\bibitem[{Ismayilzada et~al.(2025)Ismayilzada, Stevenson, and Plas}]{ismayilzada_evaluating_2025}
Mete Ismayilzada, Claire Stevenson, and Lonneke van~der Plas. 2025.
\newblock \href {https://doi.org/10.48550/arXiv.2411.02316} {Evaluating {Creative} {Short} {Story} {Generation} in {Humans} and {Large} {Language} {Models}}.
\newblock \emph{arXiv preprint}.
\newblock ArXiv:2411.02316 [cs].

\bibitem[{Kababgi et~al.(2024)Kababgi, Grisot, Pennino, and Herrmann}]{kababgi2024recognising}
Daniel Kababgi, Giulia Grisot, Federico Pennino, and J.~Berenike Herrmann. 2024.
\newblock Recognising nonnamed spatial entities in literary texts: A novel spatial entities classifier.
\newblock In \emph{Proceedings of the Computational Humanities Research Conference (CHR 2024)}, pages 472--481.

\bibitem[{Kubinec(2023)}]{kubinec_ordered_2023}
Robert Kubinec. 2023.
\newblock Ordered beta regression: A parsimonious, well-fitting model for continuous data with lower and upper bounds.
\newblock \emph{Political Analysis}, 31(4):519--536.

\bibitem[{Landis and Koch(1977)}]{landis1977}
J.~Richard Landis and Gary~G. Koch. 1977.
\newblock \href {https://doi.org/10.2307/2529310} {The measurement of observer agreement for categorical data}.
\newblock \emph{Biometrics}, 33(1):159--174.

\bibitem[{Lee and Lim(2024)}]{lee_language_2024}
Bruce~W. Lee and JaeHyuk Lim. 2024.
\newblock \href {https://doi.org/10.48550/arXiv.2402.11349} {Language {Models} {Don}'t {Learn} the {Physical} {Manifestation} of {Language}}.
\newblock \emph{arXiv preprint}.
\newblock ArXiv:2402.11349 [cs].

\bibitem[{Lenth and Piaskowski(2026)}]{lenth_emmeans_2026}
Russell~V. Lenth and Julia Piaskowski. 2026.
\newblock \href {https://rvlenth.github.io/emmeans/} {\emph{emmeans: Estimated Marginal Means, aka Least-Squares Means}}.
\newblock R package version 2.0.3.

\bibitem[{Lucy and Bamman(2021)}]{lucy_gender_2021}
Li~Lucy and David Bamman. 2021.
\newblock \href {https://doi.org/10.18653/v1/2021.nuse-1.5} {Gender and {Representation} {Bias} in {GPT}-3 {Generated} {Stories}}.
\newblock In \emph{Proceedings of the {Third} {Workshop} on {Narrative} {Understanding}}, pages 48--55, Virtual. Association for Computational Linguistics.

\bibitem[{Mahlberg(2013)}]{mahlberg_corpus_2013}
Michaela Mahlberg. 2013.
\newblock \href {https://doi.org/10.4324/9780203076088} {\emph{Corpus stylistics and {Dickens}'s fiction}}.
\newblock Routledge advances in corpus linguistics; 14. Routledge, New York.

\bibitem[{Marco et~al.(2024)Marco, Gonzalo, Mateo-Girona, and Santos}]{marco_pron_2024}
Guillermo Marco, Julio Gonzalo, M.Teresa Mateo-Girona, and Ramón Del~Castillo Santos. 2024.
\newblock \href {https://doi.org/10.18653/v1/2024.emnlp-main.1096} {Pron vs {Prompt}: {Can} {Large} {Language} {Models} already {Challenge} a {World}-{Class} {Fiction} {Author} at {Creative} {Text} {Writing}?}
\newblock In \emph{Proceedings of the 2024 {Conference} on {Empirical} {Methods} in {Natural} {Language} {Processing}}, pages 19654--19670, Miami, Florida, USA. Association for Computational Linguistics.

\bibitem[{Norris(1899)}]{norris_mcteague_1899}
Frank Norris. 1899.
\newblock \href {https://www.gutenberg.org/ebooks/165} {\emph{McTeague: A story of San Francisco}}.
\newblock Project Gutenberg.

\bibitem[{Paech(2023)}]{paech2023eqbench}
Samuel~J. Paech. 2023.
\newblock \href {https://arxiv.org/abs/2312.06281} {Eq-bench: An emotional intelligence benchmark for large language models}.
\newblock \emph{Preprint}, arXiv:2312.06281.

\bibitem[{Rohrbacher(2025{\natexlab{a}})}]{rohrbacher2025b}
Katrin Rohrbacher. 2025{\natexlab{a}}.
\newblock \href {https://doi.org/10.5334/johd.350} {de-corp: A corpus of german-language fiction and non-fiction (1780--1930)}.
\newblock \emph{Journal of Open Humanities Data}, 11(1):51.

\bibitem[{Rohrbacher(2025{\natexlab{b}})}]{rohrbacher_opening_2025a}
Katrin Rohrbacher. 2025{\natexlab{b}}.
\newblock \href {https://doi.org/10.26083/tuprints-00030149} {Opening worlds: Narrative beginnings and the role of setting}.
\newblock \emph{CCLS2025 Conference Preprints}, 4(1).

\bibitem[{Rohrbacher(forthcoming)}]{author_forthcoming}
Katrin Rohrbacher. forthcoming.
\newblock ``lived space'': A computational study of setting in fiction.
\newblock In R.~M. Aust, G.~Grisot, and B.~Herrmann, editors, \emph{Comparing landscapes: Approaches to space and affect in literary fiction}. Bielefeld University Press.

\bibitem[{Sears and Weisberg(2026)}]{sears_bot_2026}
Sydney Sears and Deena~Skolnick Weisberg. 2026.
\newblock \href {https://doi.org/10.1017/jdm.2026.10042} {Bot or not: {Can} people tell the difference between stories written by a human or by an {AI} system?}
\newblock \emph{Judgment and Decision Making}, 21:e21.

\bibitem[{Soni et~al.(2023)Soni, Sihra, Evans, Wilkens, and Bamman}]{soni-etal-2023-grounding}
Sandeep Soni, Amanpreet Sihra, Elizabeth Evans, Matthew Wilkens, and David Bamman. 2023.
\newblock \href {https://doi.org/10.18653/v1/2023.acl-long.655} {Grounding characters and places in narrative text}.
\newblock In \emph{Proceedings of the 61st Annual Meeting of the Association for Computational Linguistics (Volume 1: Long Papers)}, pages 11723--11736, Toronto, Canada. Association for Computational Linguistics.

\bibitem[{Ströker(1965)}]{stroker_philosophische_1965}
E.~Ströker. 1965.
\newblock \href {https://books.google.de/books?id=skEAAAAAMAAJ} {\emph{Philosophische {Untersuchungen} zum {Raum}}}.
\newblock Philosophische {Abhandlungen}. V. Klostermann.

\bibitem[{Thierry(1918)}]{thierry_adventure_1918}
James~Francis Thierry. 1918.
\newblock \href {https://www.gutenberg.org/ebooks/31135} {\emph{The Adventure of the Eleven Cuff-Buttons}}.
\newblock Project Gutenberg.

\bibitem[{Tian et~al.(2024)Tian, Huang, Liu, Jiang, Spangher, Chen, May, and Peng}]{tian_are_2024}
Yufei Tian, Tenghao Huang, Miri Liu, Derek Jiang, Alexander Spangher, Muhao Chen, Jonathan May, and Nanyun Peng. 2024.
\newblock \href {https://doi.org/10.48550/arXiv.2407.13248} {Are {Large} {Language} {Models} {Capable} of {Generating} {Human}-{Level} {Narratives}?}
\newblock \emph{arXiv preprint}.
\newblock ArXiv:2407.13248 [cs].

\bibitem[{Underwood et~al.(2025)Underwood, Nelson, and Wilkens}]{underwood_can_2025}
Ted Underwood, Laura~K. Nelson, and Matthew Wilkens. 2025.
\newblock \href {https://doi.org/10.48550/arXiv.2505.00030} {Can {Language} {Models} {Represent} the {Past} without {Anachronism}?}
\newblock \emph{arXiv preprint}.
\newblock ArXiv:2505.00030 [cs].

\bibitem[{Vaswani et~al.(2023)Vaswani, Shazeer, Parmar, Uszkoreit, Jones, Gomez, Kaiser, and Polosukhin}]{vaswani_attention_2023}
Ashish Vaswani, Noam Shazeer, Niki Parmar, Jakob Uszkoreit, Llion Jones, Aidan~N. Gomez, Lukasz Kaiser, and Illia Polosukhin. 2023.
\newblock \href {https://doi.org/10.48550/arXiv.1706.03762} {Attention {Is} {All} {You} {Need}}.
\newblock \emph{arXiv preprint}.
\newblock ArXiv:1706.03762 [cs].

\bibitem[{Vauth et~al.(2021)Vauth, Hatzel, Gius, and Biemann}]{vauth-etal-2021-automated}
Michael Vauth, Hans~Ole Hatzel, Evelyn Gius, and Chris Biemann. 2021.
\newblock \href {https://ceur-ws.org/Vol-2989/short_paper18.pdf} {Automated event annotation in literary texts}.
\newblock In \emph{Proceedings of the Conference on Computational Humanities Research 2021}, volume 2989 of \emph{CEUR Workshop Proceedings}, pages 333--345, Amsterdam, the Netherlands. CEUR-WS.org.

\bibitem[{Wang et~al.(2025)Wang, Hu, Li, Wang, Li, Hu, and Tan}]{wang-etal-2025-generating}
Qianyue Wang, Jinwu Hu, Zhengping Li, Yufeng Wang, Daiyuan Li, Yu~Hu, and Mingkui Tan. 2025.
\newblock \href {https://doi.org/10.18653/v1/2025.naacl-long.63} {Generating long-form story using dynamic hierarchical outlining with memory-enhancement}.
\newblock In \emph{Proceedings of the 2025 Conference of the Nations of the Americas Chapter of the Association for Computational Linguistics: Human Language Technologies (Volume 1: Long Papers)}, pages 1352--1391, Albuquerque, New Mexico. Association for Computational Linguistics.

\bibitem[{Wankmüller(2024)}]{wankmuller_introduction_2024}
Sandra Wankmüller. 2024.
\newblock \href {https://doi.org/10.1177/00491241221134527} {Introduction to {Neural} {Transfer} {Learning} {With} {Transformers} for {Social} {Science} {Text} {Analysis}}.
\newblock \emph{Sociological Methods \& Research}, 53(4):1676--1752.
\newblock Publisher: SAGE Publications Inc.

\bibitem[{Zhao et~al.(2024)Zhao, Zhang, {Wenyi Li}, Huang, Guo, Peng, Hao, Wen, Hu, Du, Guo, Li, and Chen}]{zhao_assessing_2024}
Yunpu Zhao, Rui Zhang, {Wenyi Li}, Di~Huang, Jiaming Guo, Shaohui Peng, Yifan Hao, Yuanbo Wen, Xing Hu, Zidong Du, Qi~Guo, Ling Li, and Yunji Chen. 2024.
\newblock \href {https://doi.org/10.1007/s11633-025-1546-4} {Assessing and {Understanding} {Creativity} in {Large} {Language} {Models}}.
\newblock ArXiv:2401.12491 [cs].

\bibitem[{Zhong et~al.(2024)Zhong, Gatti, Cho, and Obrist}]{zhong_exploring_2024}
Shu Zhong, Elia Gatti, Youngjun Cho, and Marianna Obrist. 2024.
\newblock \href {https://doi.org/10.48550/arXiv.2406.06587} {Exploring {Human}-{AI} {Perception} {Alignment} in {Sensory} {Experiences}: {Do} {LLMs} {Understand} {Textile} {Hand}?}
\newblock \emph{arXiv preprint}.
\newblock ArXiv:2406.06587 [cs].

\end{thebibliography}

\clearpage
\appendix

\section{English setting classifier}

\subsection{Annotation Guidelines}
\label{app:annotation}
\subsection*{General Principles}
Each sentence is assigned to exactly one category. Where multiple space types are present, assign the category that is most prominent. Sentences where space is implied but no concrete or atmospheric markers are present (e.g., references to imagined, dreamt, or remembered spaces) are labeled \textit{no space}. Only spaces that are part of the concrete storyworld are considered.
\subsection*{Category Definitions}
\paragraph{Action space (\textit{Aktionsraum})} Space that is moved through or interacted with by a character. Objects serve functional roles, enabling or hindering movement and goal-directed action. The character appropriates space through touch and bodily movement rather than through observation or sensory perception.
\textit{Example:} \textit{``He jumped up, jerked the window-shade, and dragged his chair closer to examine the shoes.''}
\paragraph{Perceived space (\textit{gestimmter Raum})} Space experienced as atmospheric and mood-laden. The character is affected by or absorbed into the environment through diffuse sensory experience — sounds, smells, light, weather — without clear directionality or goal-oriented movement. Setting may function quasi-anthropomorphically, as though it acts upon the character.
\textit{Example:} \textit{``The terror of loneliness among those overhanging mountains gripped at the boy's throat.''}
\paragraph{Visual space (\textit{Anschauungsraum})} Space observed from a static or near-static position. The character surveys the environment with their eyes rather than moving through it. Space presents itself to the character; the focus is on what is seen rather than on how the character is affected.
\textit{Example:} \textit{``He glanced from Tom to the cabin.''}
\paragraph{Descriptive space} Spatial information that situates characters or objects without being anchored to any character's perception, agency, or emotional experience. Functions as neutral scene-setting or localization.
\textit{Example:} \textit{``On either side of the towpath were farms and gardens.''}
\paragraph{No space} No spatial relationship is present, or space appears only as imagined, dreamt, remembered, or planned, i.e., not part of the concrete storyworld.
\textit{Example:} \textit{``By this curious turn I have gained the reputation of deliberate heartlessness.''}
\subsection*{Common Ambiguities}
\textbf{Action vs. perceived space.} When movement and atmosphere co-occur, assign the category that predominates. If a character moves through space but is primarily affected by or absorbed into it, prefer perceived space. If movement and goal-directed interaction with objects are foregrounded, prefer action space.
\textbf{Perceived vs. visual space.} Both involve a relatively static character, but visual space is detached and observational, while perceived space involves affective absorption. If the environment is rendered anthropomorphic or the character is emotionally moved by it, prefer perceived space.
\textbf{Descriptive vs. visual space.} Descriptive space is not anchored to any character's point of view; visual space is. If a character is explicitly observing the described scene, prefer visual space.

\subsection{Model and Training}
\label{app:eng-classifier}

The English setting classifier was fine-tuned from RoBERTa (“roberta-base”)\footnote{\url{https://huggingface.co/FacebookAI/roberta-base}}, an encoder model based on the BERT architecture \citep{devlin_bert_2019}, trained for 5 epochs with a learning rate of 1e-5 on 70\% of the annotated data and evaluated on a held-out test set of 30\% ($n = 972$). Classification performance is reported in Table~\ref{tab:english-classifier}. For comparison, Table~\ref{tab:german-classifier} reports the performance of the German classifier, adapted from \citet{rohrbacher_opening_2025a}, which achieves a macro F1 of 0.85, broadly similar to the English classifier, with the German classifier performing slightly higher overall.

\begin{table}[h]
\centering
\small
\begin{tabular}{lccc}
\toprule
\textbf{Class} & \textbf{Precision} & \textbf{Recall} & \textbf{F1-score} \\
\midrule
Perceived space   & 0.81 & 0.82 & 0.82 \\
Action space      & 0.87 & 0.83 & 0.85 \\
Visual space      & 0.65 & 0.76 & 0.70 \\
Descriptive space & 0.75 & 0.85 & 0.80 \\
No space          & 0.95 & 0.90 & 0.92 \\
\midrule
Macro avg         & 0.81 & 0.83 & 0.82 \\
\bottomrule
\end{tabular}
\caption{Classification report for the English setting classifier. Precision, Recall, and F1-score are reported per class, along with macro averages.}
\label{tab:english-classifier}
\end{table}

\begin{table}[h]
\centering
\small
\begin{tabular}{lccc}
\toprule
\textbf{Class} & \textbf{Precision} & \textbf{Recall} & \textbf{F1} \\
\midrule
Perceived space   & 0.86 & 0.82 & 0.84 \\
Action space      & 0.91 & 0.81 & 0.86 \\
Visual space      & 0.79 & 0.88 & 0.83 \\
Descriptive space & 0.78 & 0.84 & 0.81 \\
No space          & 0.86 & 0.93 & 0.91 \\
\midrule
Macro avg         & 0.84 & 0.86 & 0.85 \\
\bottomrule
\end{tabular}
\caption{Classification report for the German setting classifier. Precision, 
Recall, and F1-score are reported per class, along with macro averages. 
Adapted from \citet{rohrbacher_opening_2025a}.}
\label{tab:german-classifier}
\end{table}

\subsection{Classifier Validation}
\label{app:clf-validation}

Since the classifier was fine-tuned on human-authored fictional prose, a 
key question is whether it generalizes to AI-generated text, which may 
differ systematically from its training domain in vocabulary or stylistic conventions. To assess this, two annotators manually labeled a stratified sample of 600 sentences drawn from the AI-generated texts used in the main study, covering all four models and both languages. Inter-annotator agreement was substantial ($\kappa = 0.736$, 79.2\% raw agreement), and remaining disagreements were resolved through discussion, producing a gold standard against which classifier performance was evaluated.

Table~\ref{tab:clf-perclass} reports per-class precision, recall, and F1 
for the classifier evaluated against the gold standard annotations. Table~\ref{tab:clf-bymodel} shows performance broken down by model and language. Results are consistent across most model-language combinations, where accuracy ranges from 77.3\% to 85.3\% with $\kappa$ values indicating substantial agreement. This confirms that the classifier generalizes well to AI-generated text. 
Mistral~3.2 shows somewhat lower performance in both 
languages, most pronounced in German (62.7\%, $\kappa = 0.533$). During 
annotation, we observed that Mistral~3.2 occasionally produced grammatically 
malformed or semantically incoherent sentences that superficially resembled 
a spatial category but could not be assigned one meaningfully. These were 
labeled as \textit{no space} by annotators. The classifier, trained on well-formed 
prose, was not exposed to such cases during fine-tuning and therefore tends 
to assign a spatial label rather than \textit{no space} in these instances, 
which disproportionately affects Mistral~3.2 across both languages and is most pronounced in German where such outputs were most frequent. Performance for all other models and languages falls within a narrow and acceptable range, supporting the classifier's general applicability to AI-generated text.

\begin{table}[h]
\centering
\small
\begin{tabular}{lcccc}
\hline
\textbf{Category} & \textbf{P} & \textbf{R} & \textbf{F1} & \textbf{n} \\
\hline
Action space      & 0.758 & 0.805 & 0.781 & 113 \\
Descriptive space & 0.733 & 0.946 & 0.826 &  93 \\
No space          & 0.933 & 0.586 & 0.720 & 191 \\
Perceived space   & 0.708 & 0.867 & 0.780 &  98 \\
Visual space      & 0.767 & 0.876 & 0.818 & 105 \\
\hline
Macro avg         & 0.780 & 0.816 & 0.785 & 600 \\
\hline
\end{tabular}
\caption{Per-class classifier performance against gold standard annotations (n=600).}
\label{tab:clf-perclass}
\end{table}

\begin{table}[h]
\centering
\small
\begin{tabular}{llccc}
\hline
\textbf{Language} & \textbf{Model} & \textbf{Accuracy} & \textbf{$\kappa$} & \textbf{Macro F1} \\
\hline
English & GPT~4.1           & 81.3\% & 0.767 & 0.810 \\
        & Gemma~3         & 77.3\% & 0.717 & 0.777 \\
        & LlaMA~3.3         & 77.3\% & 0.717 & 0.773 \\
        & Mistral~3.2       & 72.0\% & 0.650 & 0.722 \\
\hline
German  & GPT~4.1          & 85.3\% & 0.817 & 0.855 \\
        & Gemma~3         & 85.3\% & 0.817 & 0.855 \\
        & LlaMA~3.3          & 82.7\% & 0.783 & 0.827 \\
        & Mistral~3.2       & 62.7\% & 0.533 & 0.645 \\
\hline
\end{tabular}
\caption{Classifier performance by model and language.}
\label{tab:clf-bymodel}
\end{table}

\subsubsection{Inter-annotator agreement}  

\begin{table}[htbp]
  \centering
  \begin{tabular}{lccr}
    \toprule
    Category          & IoU          & $\kappa$ & $n_{\cup}$ \\
    \midrule
    Action space      & 0.672        & 0.761 & 128 \\
    Descriptive space & 0.734        & 0.818 & 109 \\
    No space          & 0.710        & 0.760 & 207 \\
    Perceived space   & 0.543        & 0.646 & 127 \\
    Visual space      & 0.604        & 0.689 & 154 \\
    \midrule
    Overall           & 79.2\,\% raw & 0.736 & 600 \\
    \bottomrule
  \end{tabular}
  \caption{Inter-annotator agreement per category on the double-coded sample ($n=600$ sentences). IoU is the number of sentences both annotators assigned to a category divided by the number $n_{\cup}$ that at least one annotator did; $\kappa$ is the chance-corrected one-vs-rest Cohen's $\kappa$. The $n_{\cup}$ column sums to 725 rather than 600 because each of the 125 disagreed sentences counts toward two categories. The underlying marginals are given in Figure~\ref{fig:confusion}\subref{fig:cm-iaa}.}
  \label{tab:iaa-per-category}
\end{table}

Table~\ref{tab:iaa-per-category} reports two per-category agreement statistics on the pooled English and German sample. IoU is a stricter variant of the standard positive-specific agreement statistic (inter-annotator F1; \citealp{cicchetti1990,hripcsak2005}), which ranks the categories identically. Perceived space shows the lowest raw agreement (IoU $=$ 0.543) but a chance-corrected $\kappa$ of 0.646, conventionally substantial agreement \citep{landis1977}. These are initial assessment values. All disagreements were subsequently
resolved through discussion, and the classifier was evaluated against the resulting adjudicated gold standard, on which its perceived-space F1 (0.780) is comparable to the other categories (0.720--0.826; Table~\ref{tab:clf-perclass}).


\begin{figure}[tbp]
  \centering
  \begin{subfigure}[b]{0.9\linewidth}
    \includegraphics[width=\linewidth]{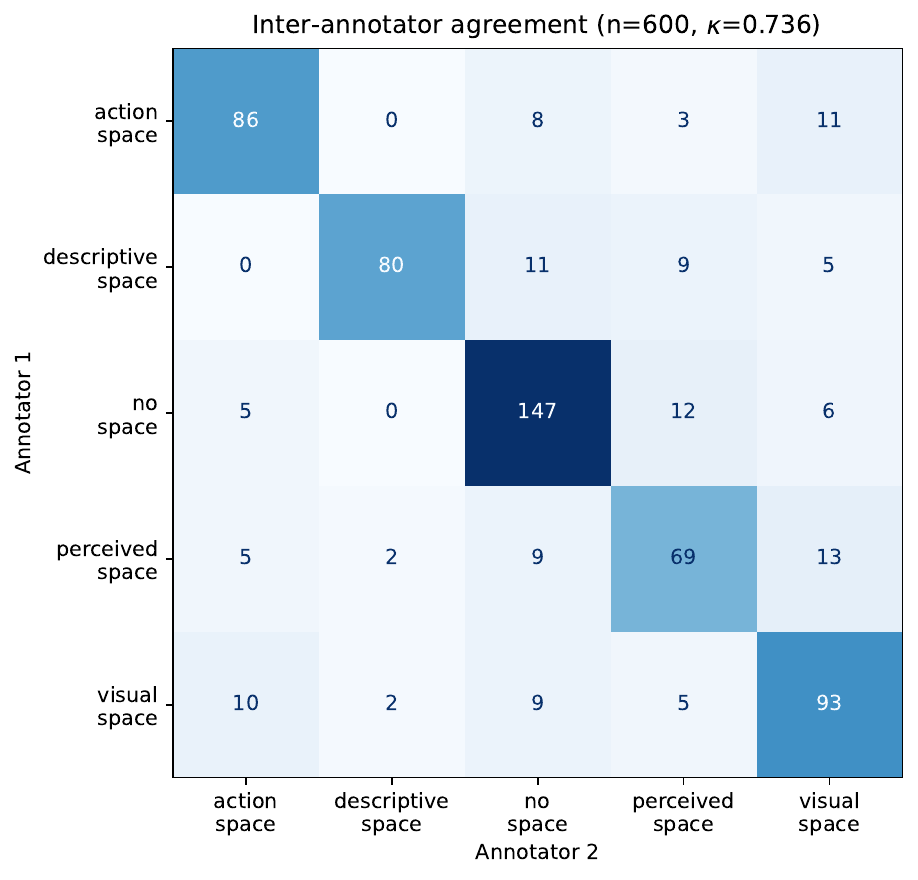}
     \caption{Annotator 1 (rows) against Annotator 2 (columns) on the
             double-coded sample, $n=600$.}
    \label{fig:cm-iaa}
  \end{subfigure}

  \vspace{1.2em}

  \begin{subfigure}[b]{\linewidth}
    \includegraphics[width=\linewidth]{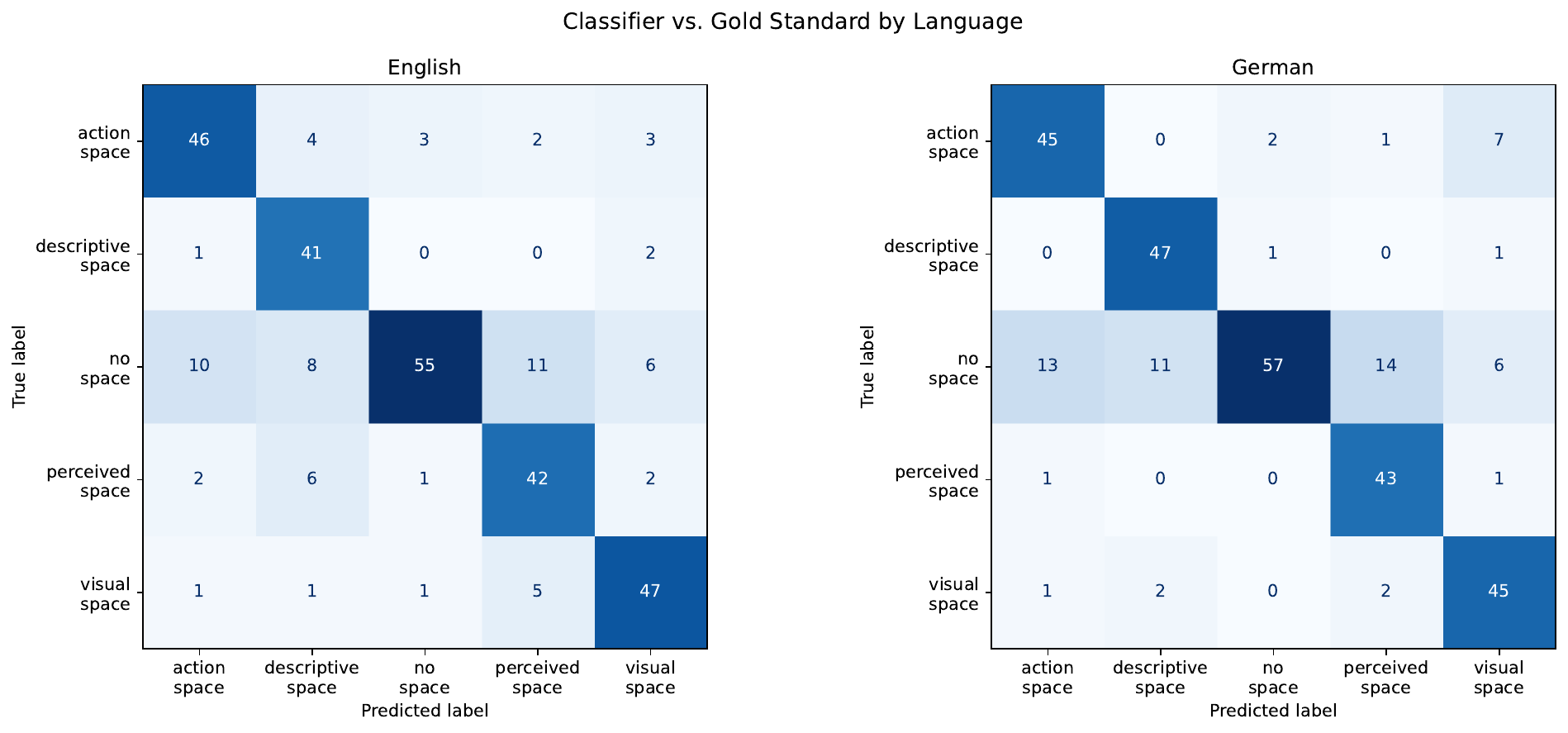}
    \caption{Classifier predictions (columns) against the adjudicated gold
             standard (rows), $n=300$ per language.}
    \label{fig:cm-clf}
  \end{subfigure}

  \caption{Confusion structure of the five-category scheme. Panel \subref{fig:cm-iaa} reports human--human agreement; panel
  \subref{fig:cm-clf} reports classifier--gold agreement.}
  \label{fig:confusion}
\end{figure}

Figure~\ref{fig:confusion} shows the confusion structure behind these agreement figures, which suggests that the disagreements follow a systematic pattern. In panel~\subref{fig:cm-iaa}, perceived space is confused principally with no space (21 sentences) and with visual space (18). Both confusions
involve threshold judgments, one about whether atmospheric content is concretely spatial at all, the other about affective absorption as against detached observation (Appendix~\ref{app:annotation}). Perceived and action space, the two categories carrying the central contrast of this paper, are almost never confused. The count is 8 of 600 sentences (1.3\%) between annotators, and 4 of
300 (English) and 2 of 300 (German) for the classifier against the gold standard. The reported difference is therefore unlikely to be an artefact of annotator or classifier uncertainty between these two categories. Panel~\subref{fig:cm-clf} shows that the classifier's errors are concentrated in the no space row. Recall for that category is 0.611 in English (55 of 90) and 0.564 in German (57 of 101). Misassigned sentences go most often to perceived space, and over half of these come from Mistral~3.2, whose outputs are frequently malformed (Section~\ref{app:clf-validation}). Only four occur across all 150 GPT~4.1 sentences. Because gold no-space sentences are by construction those with the weakest spatial signal, this asymmetry inflates the measured amount of represented space, but is concentrated in Mistral~3.2, whose deviations are among the smallest of the four models reported in Section~\ref{sec:temporal}. The central contrast is unaffected.

\section{Deviation from Human Baseline: English and German}
\label{app:ger-deviation}

\begin{figure*}[h]
    \centering
    \includegraphics[width=\textwidth]{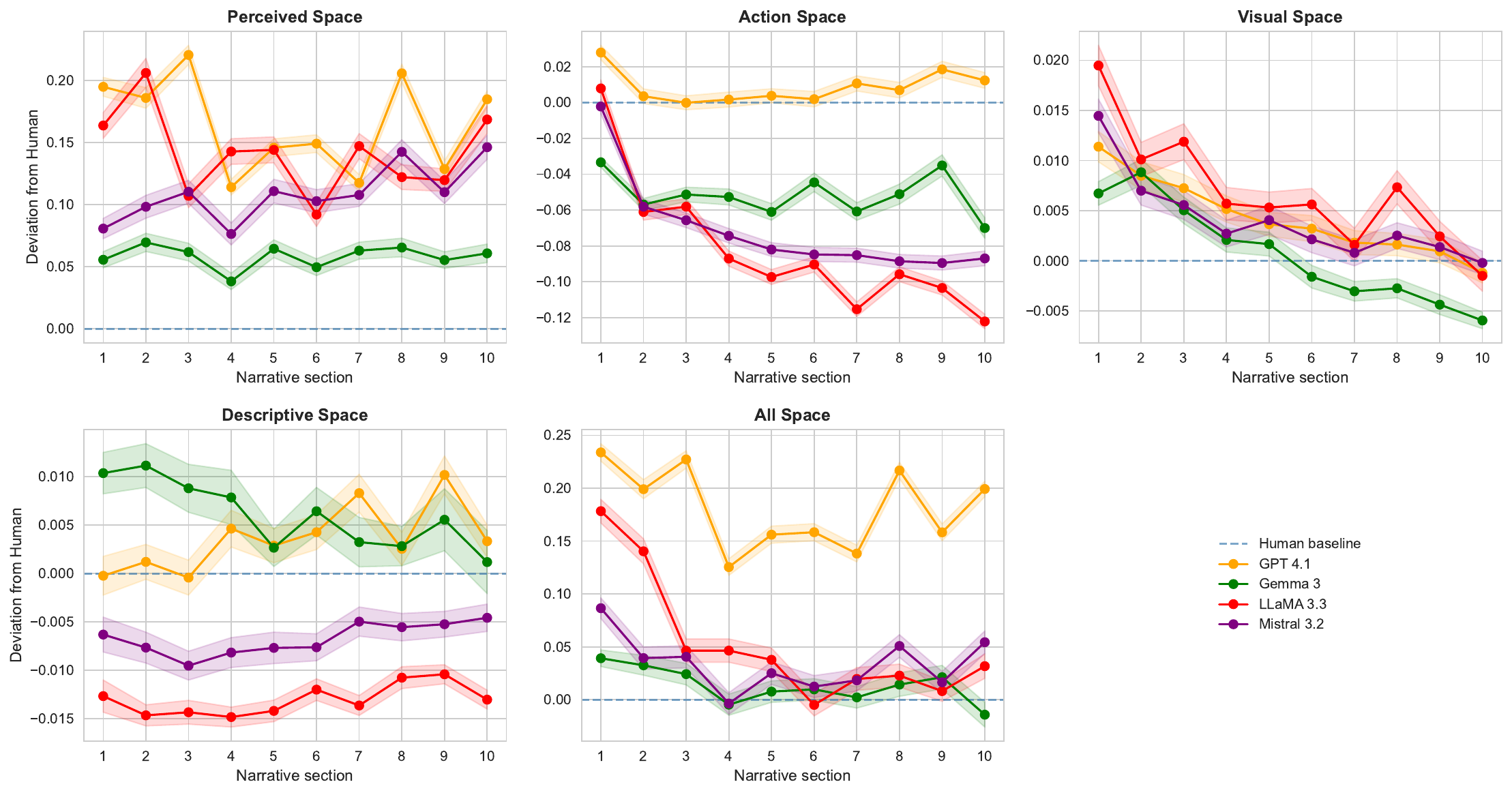}
    \vspace{1em}
    \includegraphics[width=\textwidth]{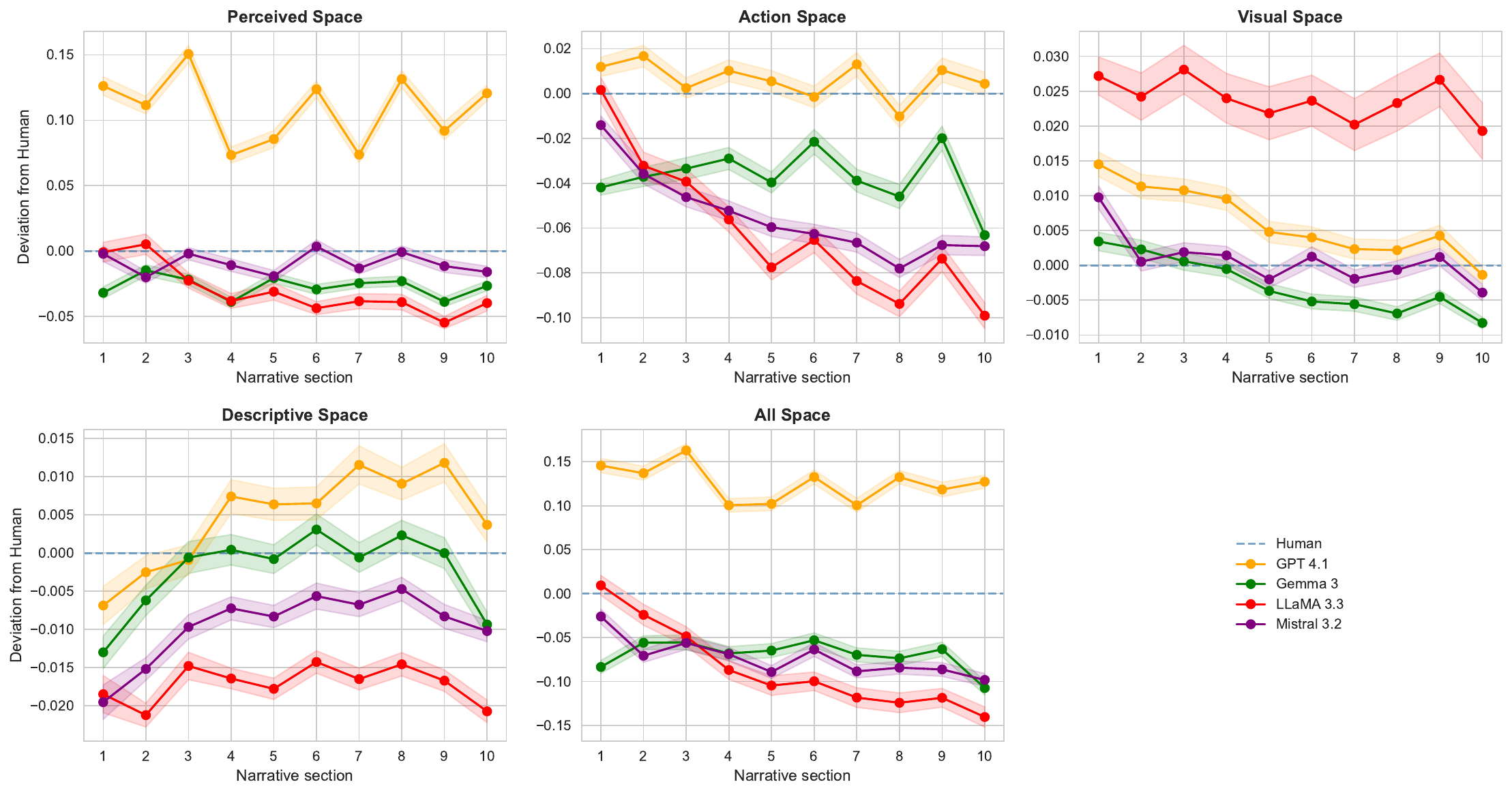}
    \caption{Deviation of AI model predictions from the human baseline 
    across narrative sections for each spatial category, for English (top) 
    and German (bottom). The human baseline represents the average normalized 
    frequency of each space type per narrative section (deviation = 0); 
    positive values indicate higher proportions than human-authored texts, 
    negative values indicate lower. Error bars indicate 95\% confidence intervals.}
    \label{fig:deviation-both}
\end{figure*}

Figure~\ref{fig:deviation-both} shows the deviation of each model 
from the human baseline across ten narrative sections for English and German.

\section{Prompt Sensitivity}
\label{app:ablation}

\begin{figure}[htbp]
    \centering
    \begin{subfigure}[t]{0.48\textwidth}
        \includegraphics[width=\textwidth]{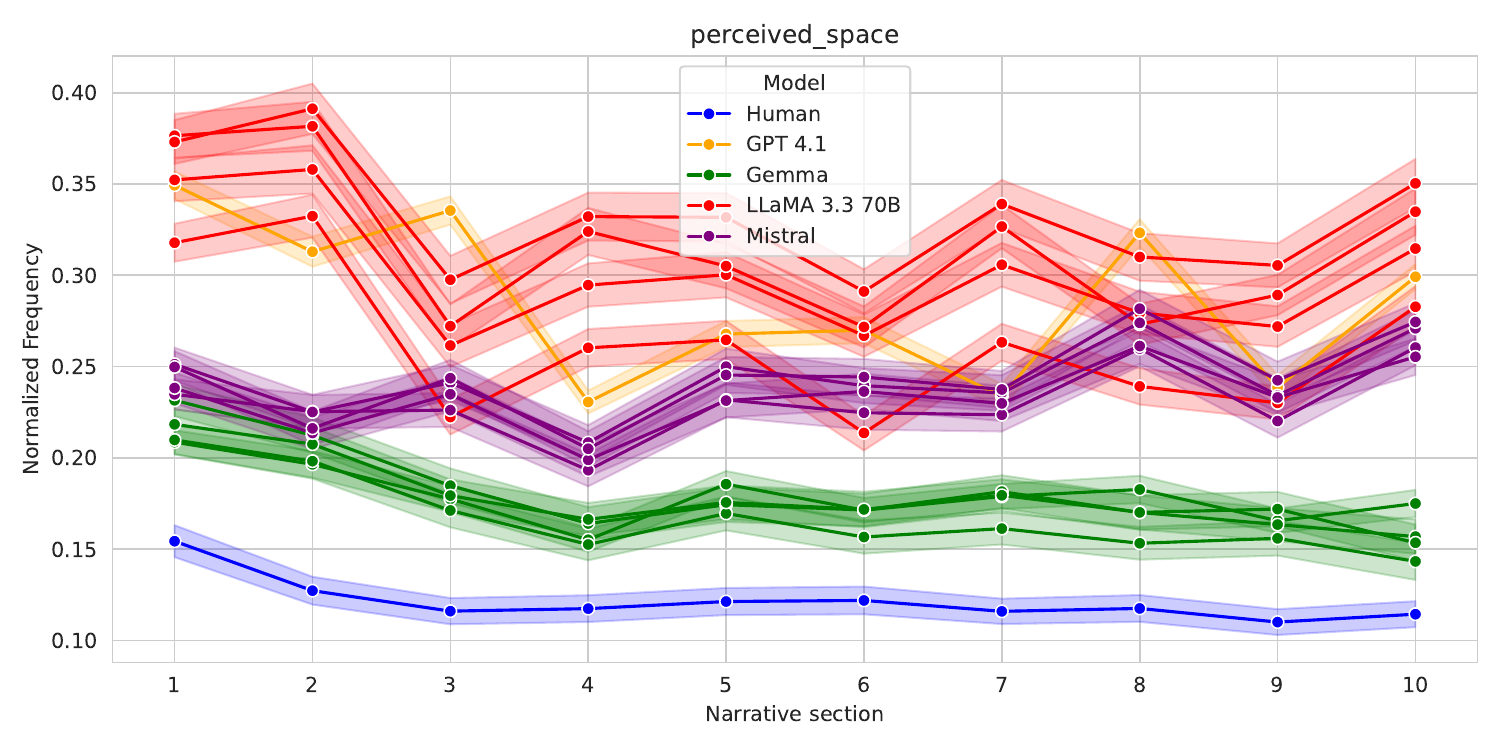}
    \end{subfigure}
    \hfill
    \begin{subfigure}[t]{0.48\textwidth}
        \includegraphics[width=\textwidth]{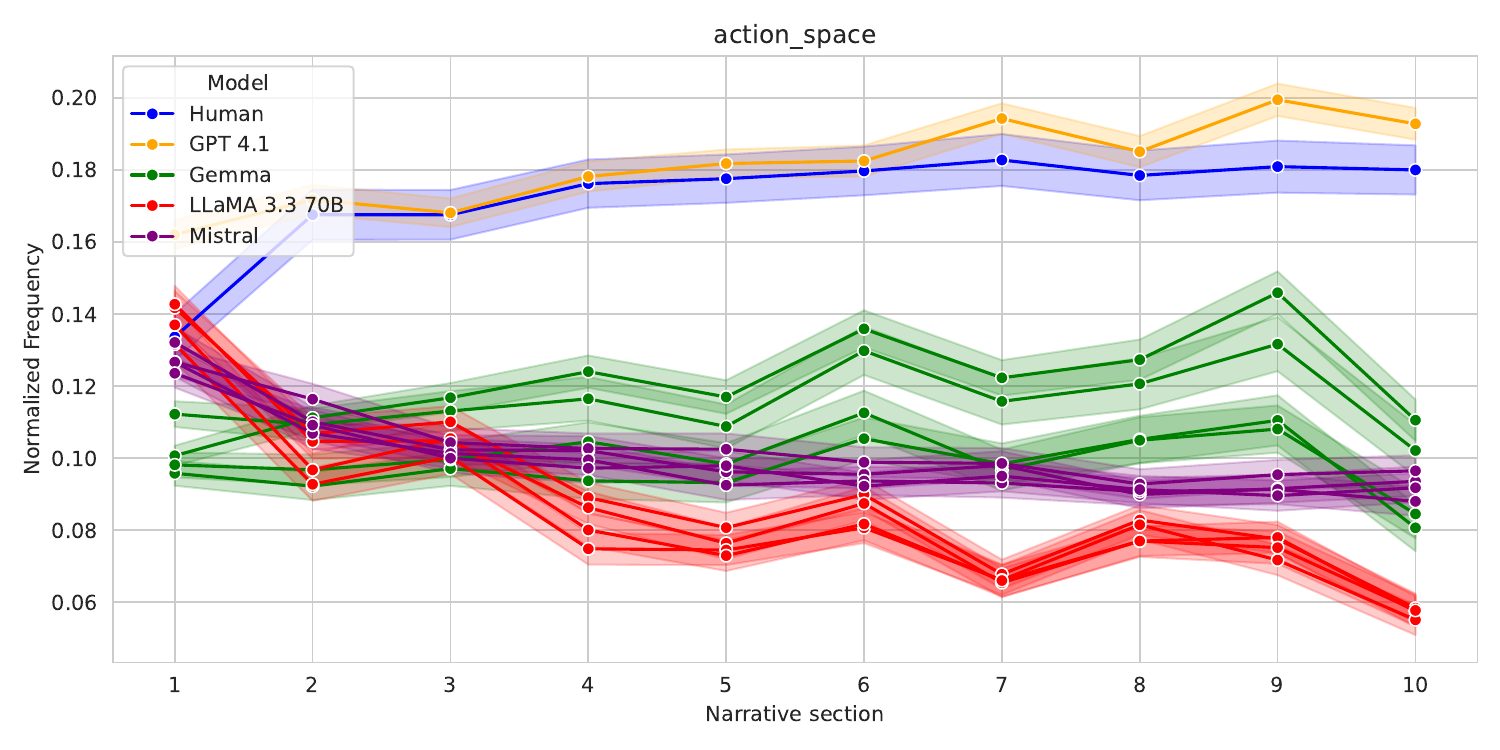}
    \end{subfigure}

    \vspace{0.5em}

    \begin{subfigure}[t]{0.48\textwidth}
        \includegraphics[width=\textwidth]{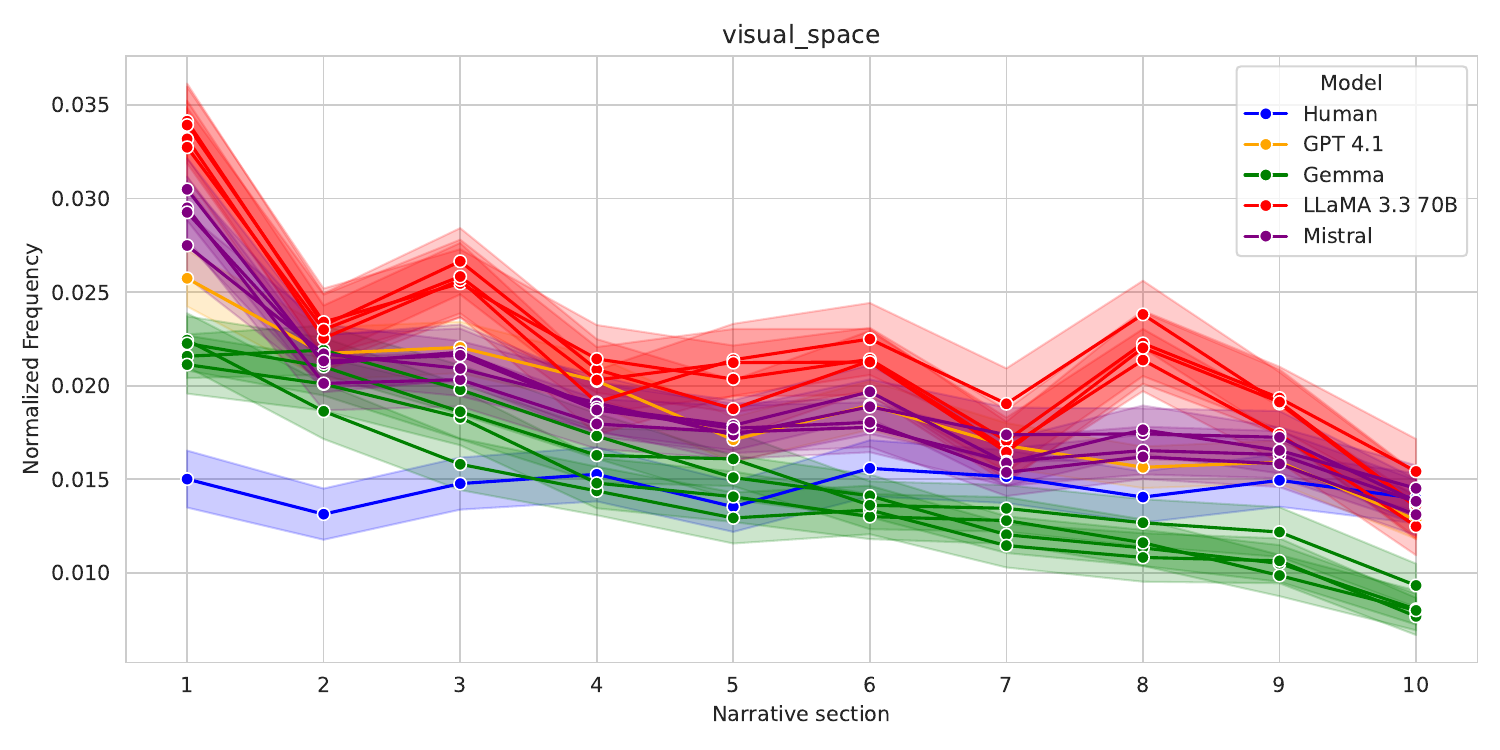}
    \end{subfigure}
    \hfill
    \begin{subfigure}[t]{0.48\textwidth}
        \includegraphics[width=\textwidth]{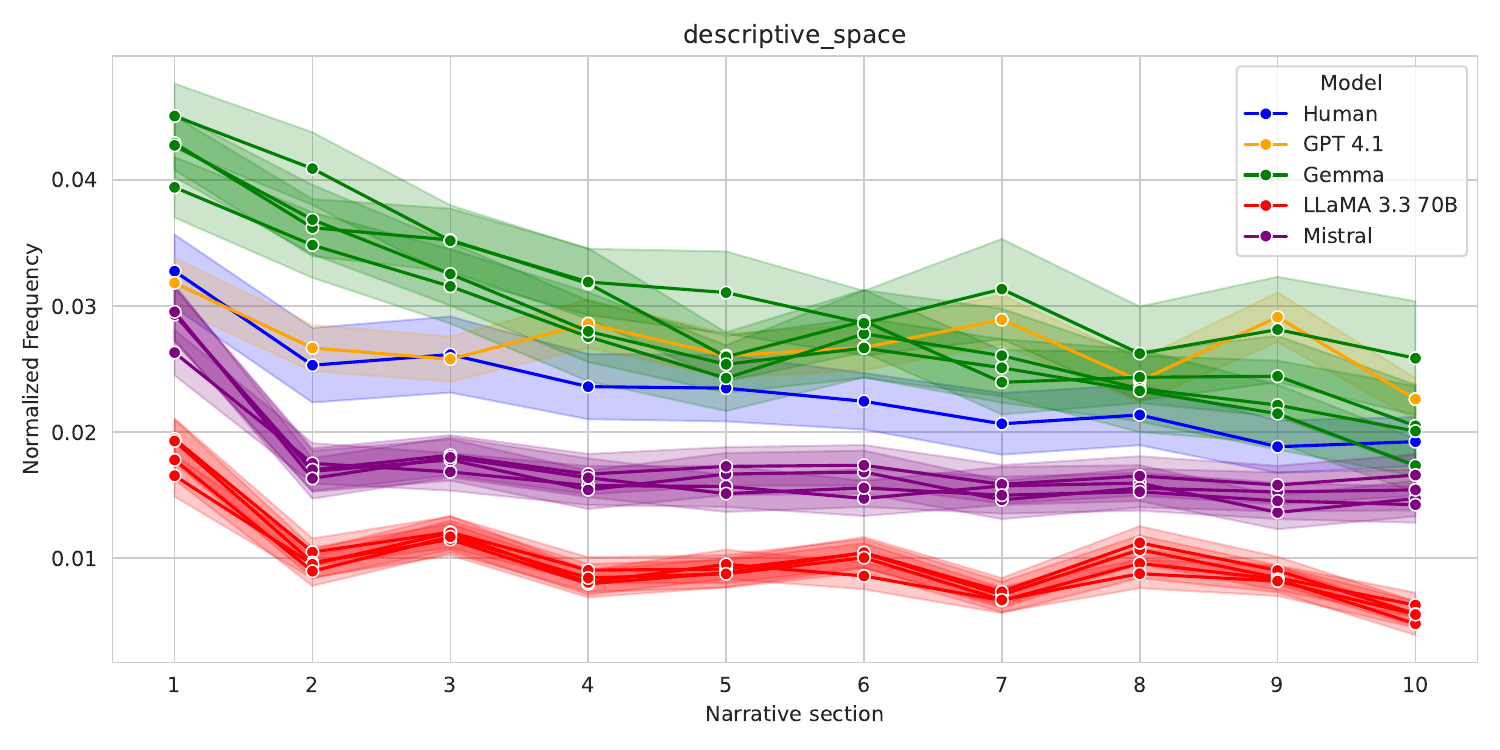}
    \end{subfigure}

    \vspace{0.5em}

    \begin{subfigure}[t]{0.48\textwidth}
        \includegraphics[width=\textwidth]{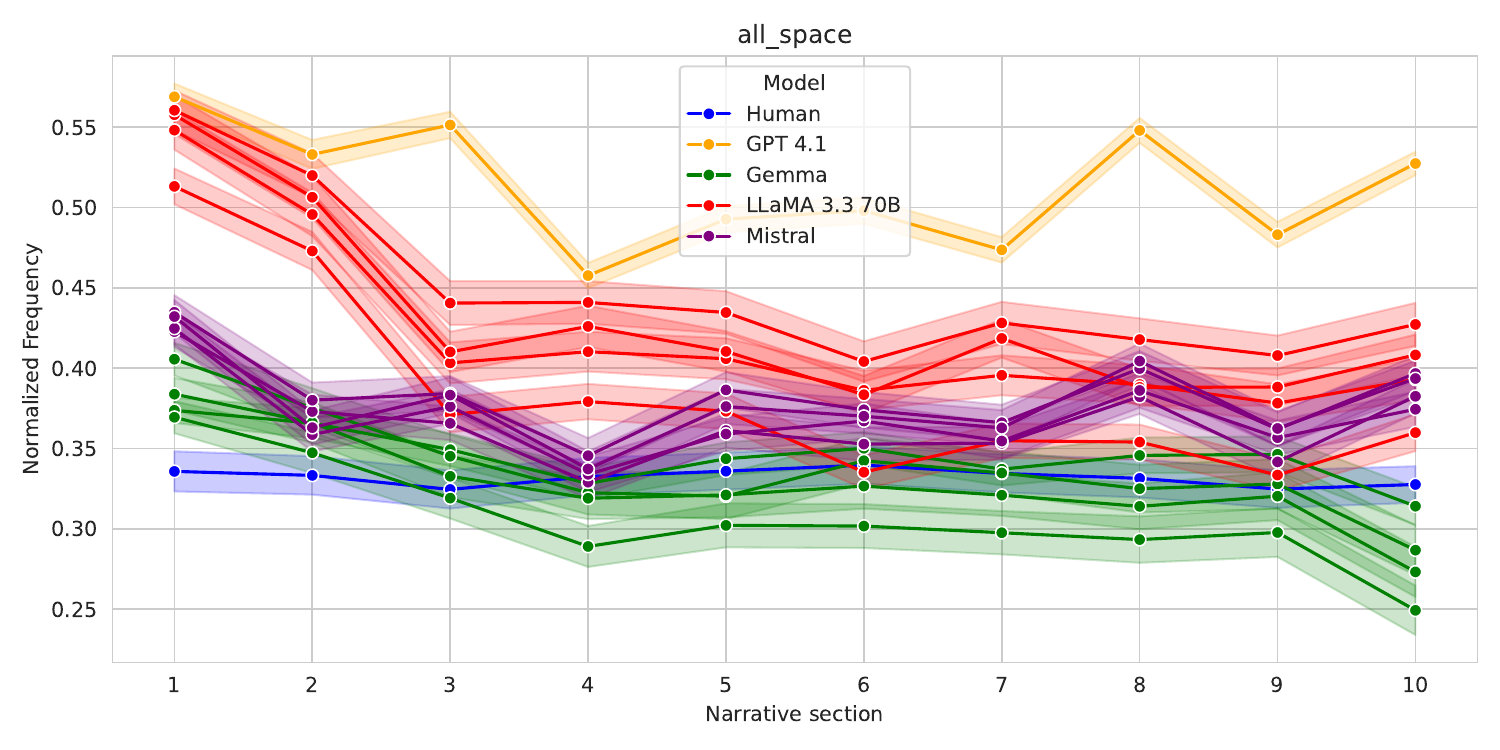}
    \end{subfigure}

    \caption{Normalized frequency of spatial categories across ten narrative 
    sections (English). Human-authored fiction is shown as a single line; 
    for each model, four lines represent the original prompt and three 
    prompt variants.}
    \label{fig:ablation-sections-en}
\end{figure}

\begin{figure}[htbp]
    \centering
    \begin{subfigure}[t]{0.48\textwidth}
        \includegraphics[width=\textwidth]{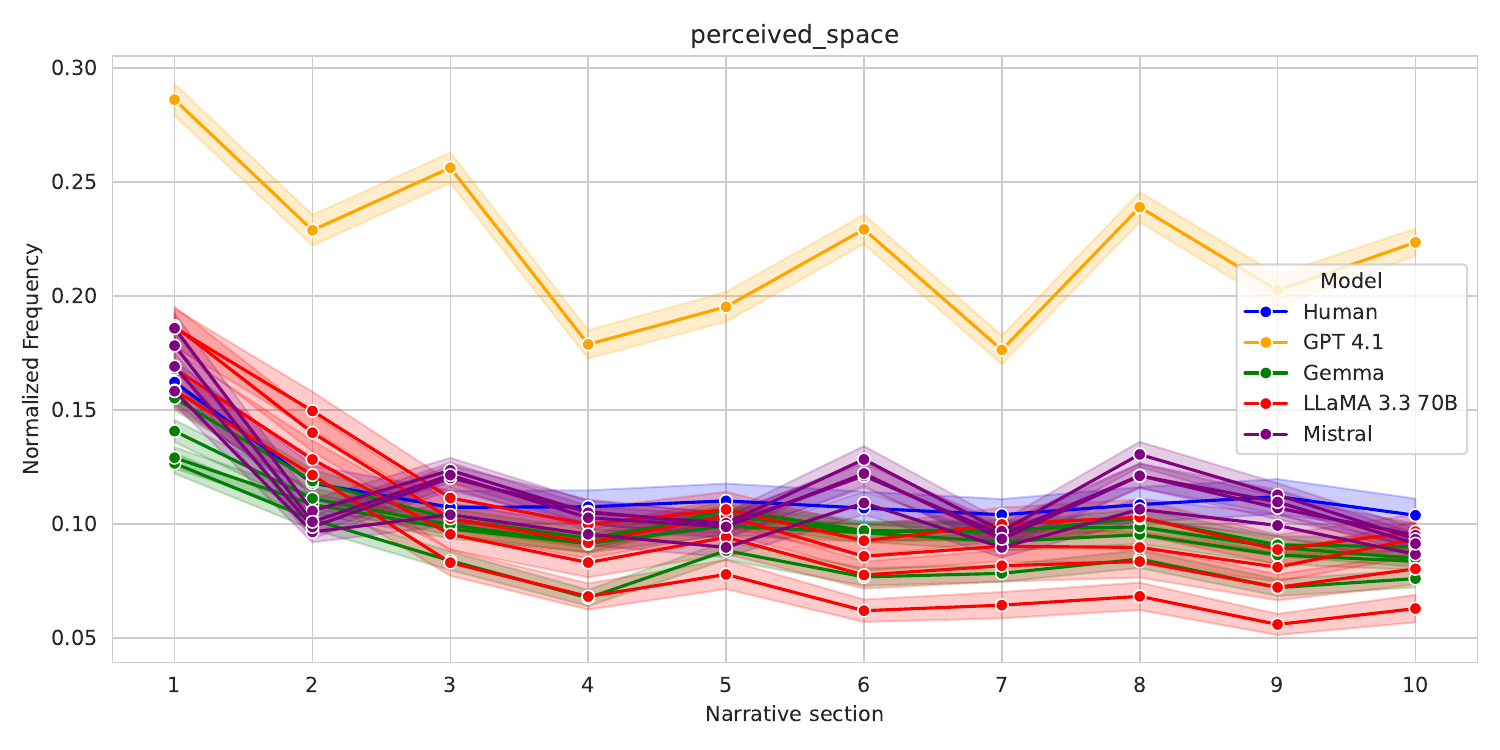}
    \end{subfigure}
    \hfill
    \begin{subfigure}[t]{0.48\textwidth}
        \includegraphics[width=\textwidth]{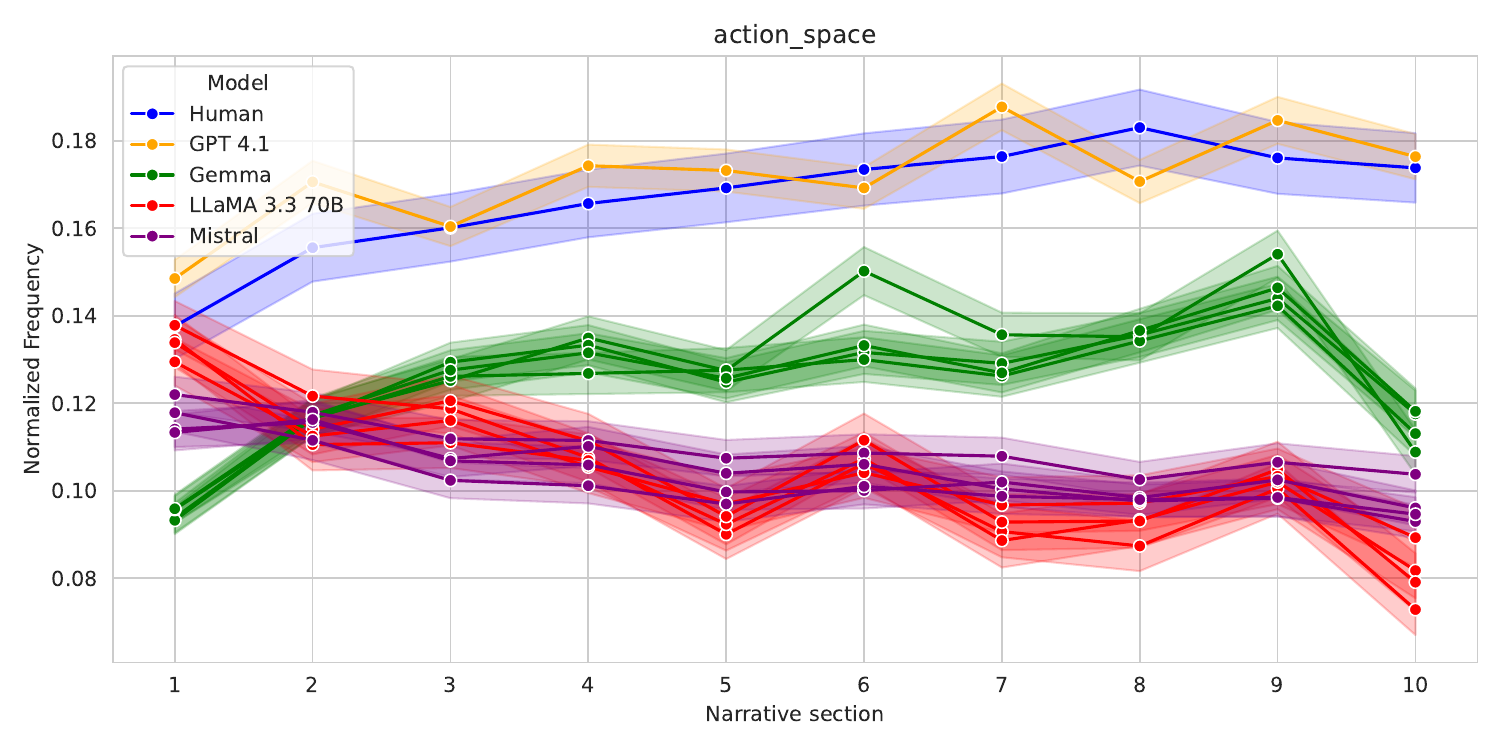}
    \end{subfigure}

    \vspace{0.5em}

    \begin{subfigure}[t]{0.48\textwidth}
        \includegraphics[width=\textwidth]{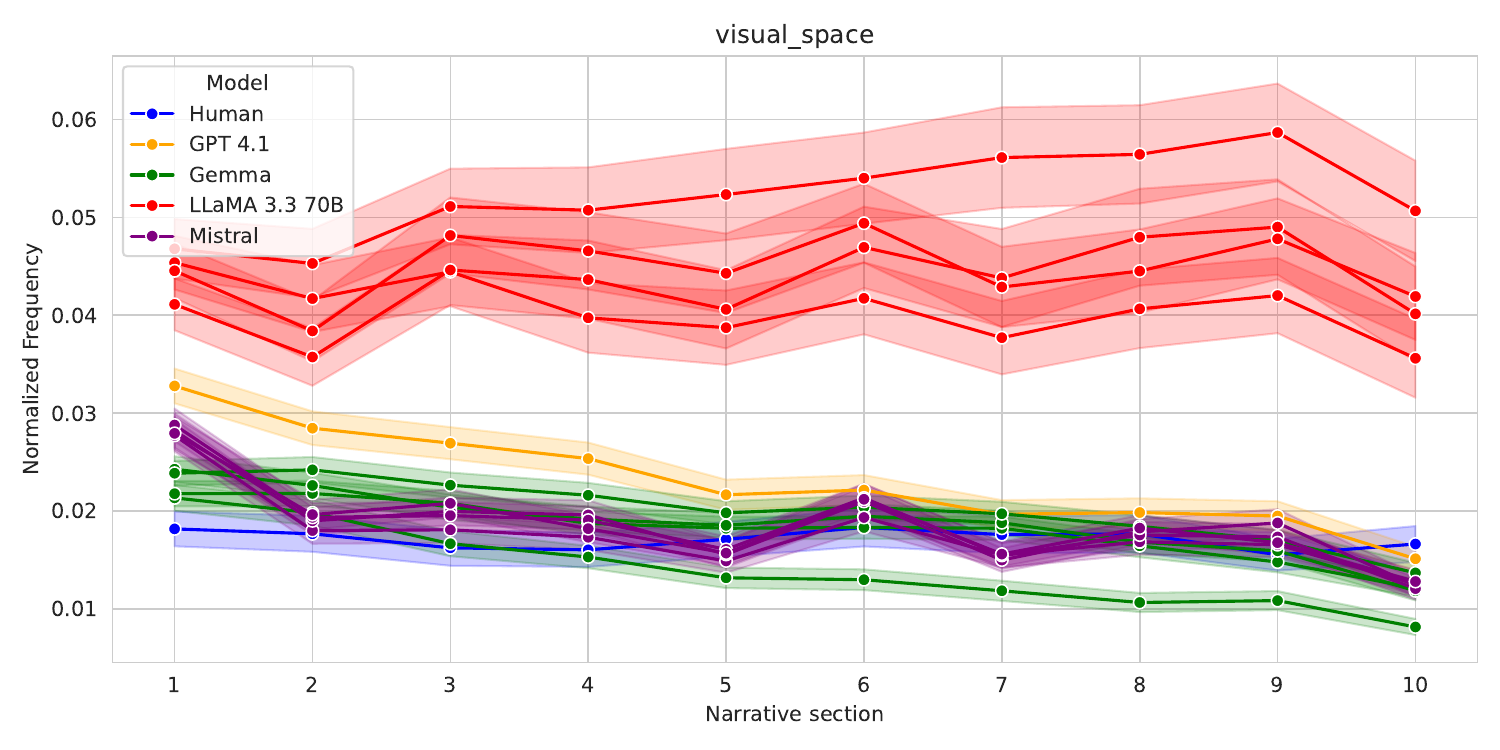}
    \end{subfigure}
    \hfill
    \begin{subfigure}[t]{0.48\textwidth}
        \includegraphics[width=\textwidth]{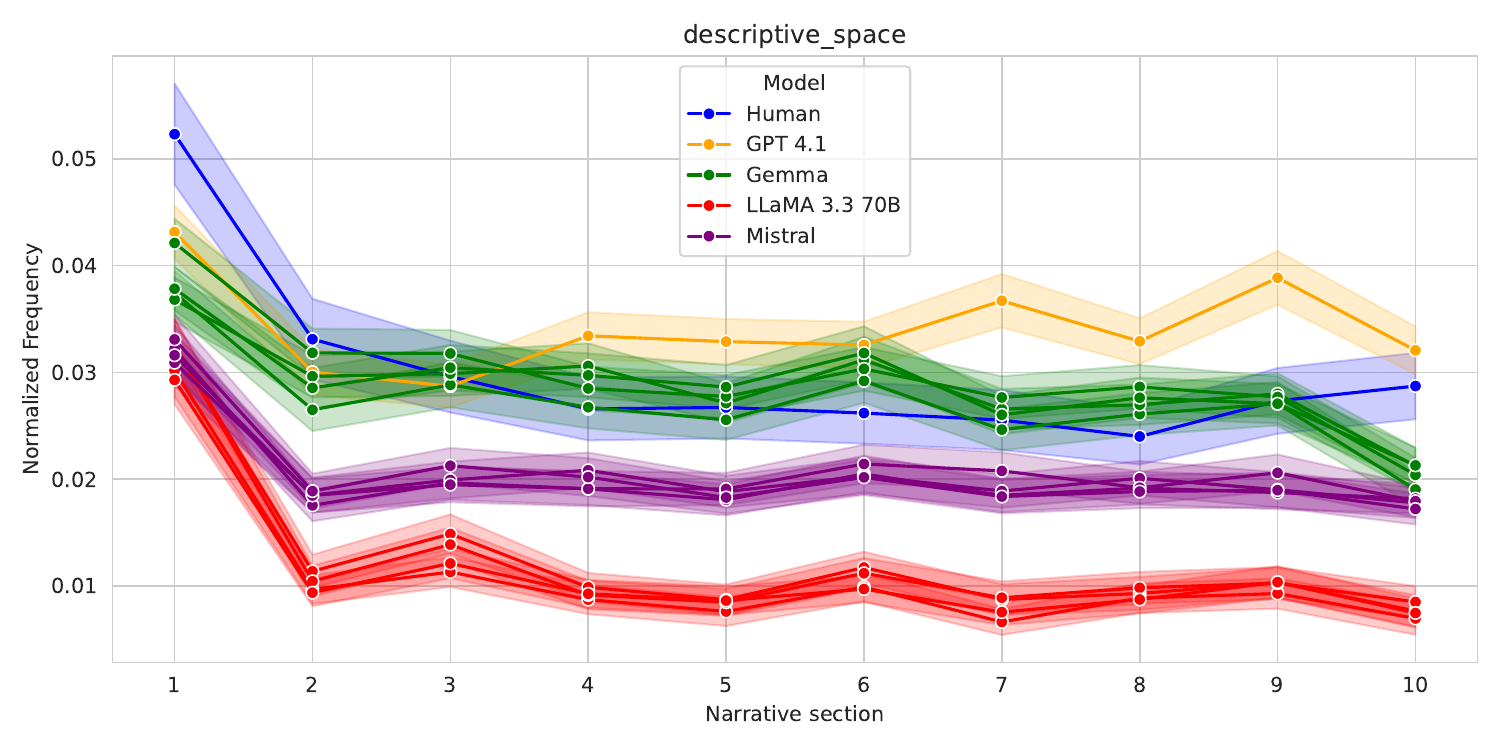}
    \end{subfigure}

    \vspace{0.5em}

    \begin{subfigure}[t]{0.48\textwidth}
        \includegraphics[width=\textwidth]{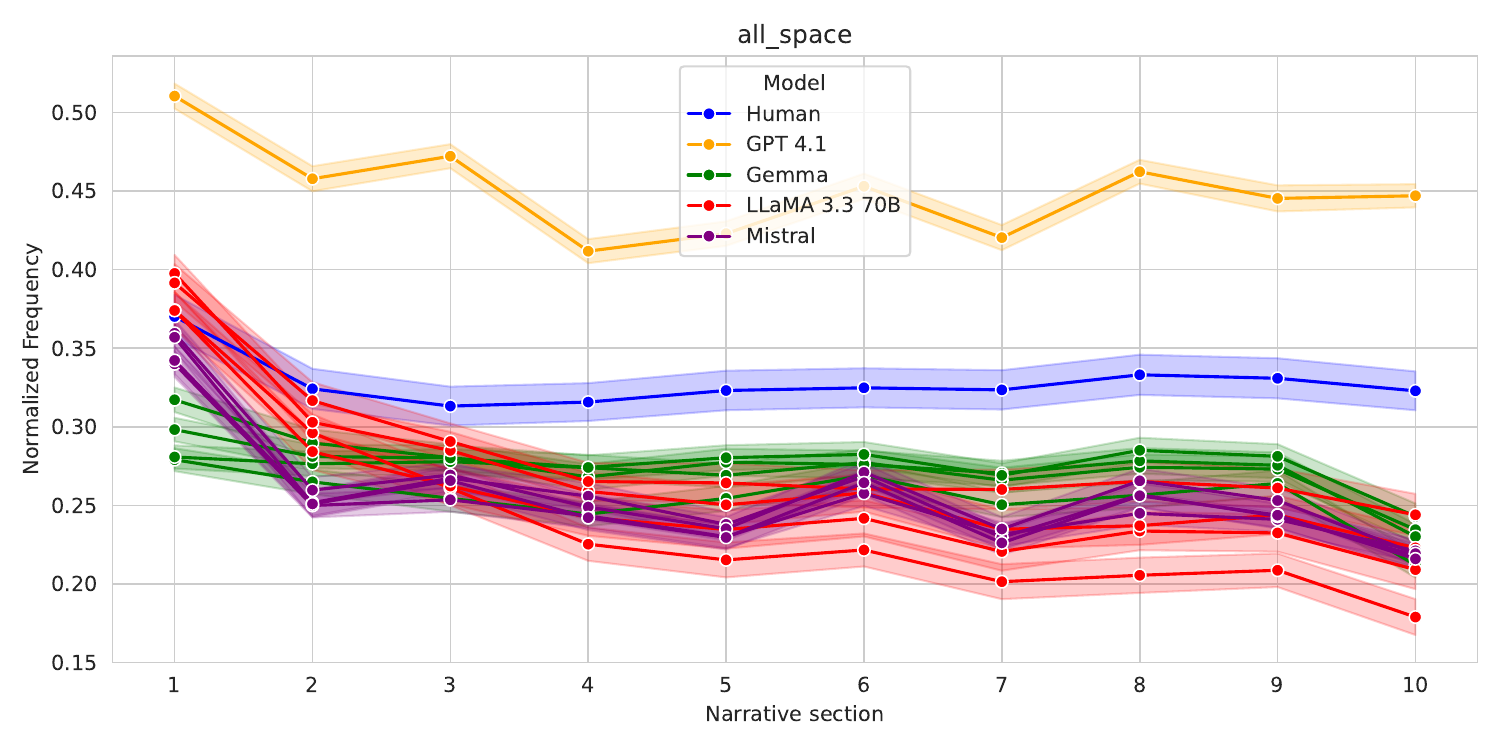}
    \end{subfigure}

    \caption{Normalized frequency of spatial categories across ten narrative 
    sections (German). Human-authored fiction is shown as a single line.
    For each model, four lines represent the original prompt and three 
    prompt variants.}
    \label{fig:ablation-sections-ger}
\end{figure}

To validate that the findings are not sensitive to prompt formulation, we ran three prompt variants (see Table~\ref{tab:ablation-prompts}) per open-source model and language in addition to the original prompt. Figures~\ref{fig:ablation-sections-en} and~\ref{fig:ablation-sections-ger} show the normalized frequency of each spatial category across ten narrative sections for the original prompt and 
three variants, for English and German respectively. Although prompt formulation introduces some variation in the 
absolute frequency of each category, the overall trends across the narrative 
sections remain consistent across all conditions. This suggests that the 
patterns reported in the main analysis are not an artifact of the specific 
prompt used.

\begin{table*}[ht]
\centering
\small
\renewcommand{\arraystretch}{1.4}
\begin{tabularx}{\textwidth}{>{\raggedright\arraybackslash}p{1.5cm}>{\raggedright\arraybackslash}p{2cm}>{\raggedright\arraybackslash}X}
\toprule
\textbf{Variant} & \textbf{Role} & \textbf{Instruction} \\
\midrule

\multirow{3}{*}{V1} 
& System & As an award-winning author, you are tasked with writing a fictional narrative. Write a continuous narrative without any interruptions or questions. Do not break the role of the author. Do not finish the story or begin to summarize the narrative. Do not repeat yourself or reuse content from previous chapters. Each chapter should continue the story by introducing new elements, deepening existing characters, and progressing the plot. \\
\cmidrule{2-3}
& Step 1 & Write a fictional story in the genre \texttt{\{genre\}}, beginning with the following opening: \texttt{\{sentence\}}. Produce the first chapter as a continuous text of approximately 3,000 tokens. Do not include any AI comments, notes, or meta-statements that would break the role of the author. \\
\cmidrule{2-3}
& Step 2+ & Write the next chapter of your narrative, continuing exactly from where the previous chapter ended. Introduce new elements, further develop the existing characters, and expand the narrative ($\sim$3,000 tokens). Do not conclude the story. Avoid repeating content from the previous chapter and do not include any AI comments or meta-statements. \\
\midrule

\multirow{3}{*}{V2} 
& System & Take the role of an award-winning author. Your task is to write a fictional narrative. Avoid interacting with the user by making meta-comments or by asking questions. Write a continuous narrative. Do not finish your story or summarize the plot; instead each chapter should introduce new ideas and build upon the established story line and characters. Let the story run as long as possible. \\
\cmidrule{2-3}
& Step 1 & Write a fictional story in the genre \texttt{\{genre\}}. Begin your narrative with the following opening sentence: \texttt{\{sentence\}}. Produce the first chapter as a continuous piece of text of around 3,000 tokens. Keep the story flowing without stopping. Avoid any meta-comments, notes, or statements that break the role of the author. \\
\cmidrule{2-3}
& Step 2+ & Write the next chapter of your narrative. Continue exactly where the previous chapter ended. Introduce new elements, further develop the existing characters, and expand the narrative ($\sim$3,000 tokens). Do not bring the story to an end. Avoid repeating content from the previous chapter and do not include any AI comments or meta-statements. \\
\midrule

\multirow{3}{*}{V3} 
& System & Assume the role of an award-winning author and compose a fictional narrative. Do not interact with the user; avoid posing questions, making meta-comments, or producing any statements outside the narrative. The story must not be concluded or summarized. Each chapter should introduce new elements and further develop the existing storyline and characters. Allow the narrative to continue for as long as possible. \\
\cmidrule{2-3}
& Step 1 & Compose a fictional narrative in the genre \texttt{\{genre\}}. Begin with the following opening sentence: \texttt{\{sentence\}}. Produce the first chapter as a continuous text of approximately 3,000 tokens, maintaining an uninterrupted narrative flow. Do not include any meta-comments, notes, or statements that would break the role of the author. \\
\cmidrule{2-3}
& Step 2+ & Compose the next chapter of the narrative, continuing exactly from where the previous chapter ended. Introduce new elements, further develop the existing characters, and expand the narrative ($\sim$3,000 tokens). Do not conclude the story. Avoid repeating material from the previous chapter and do not include any AI comments or meta-statements. \\

\bottomrule
\end{tabularx}
\caption{Prompt variants used in the prompt sensitivity ablation. V1--V3 vary in phrasing and verbosity. \texttt{\{genre\}} and \texttt{\{sentence\}} are filled with metadata and the opening sentence of each source text. Step~2 is repeated for each subsequent chapter.}
\label{tab:ablation-prompts}
\end{table*}

\section{Temporal Distribution of Setting across Narrative Time}
\label{app:temporal-dist}

\begin{figure*}[h]
\centering
\begin{minipage}[t]{0.48\textwidth}
    \centering
    \begin{subfigure}[t]{0.48\linewidth}
        \includegraphics[width=\linewidth]{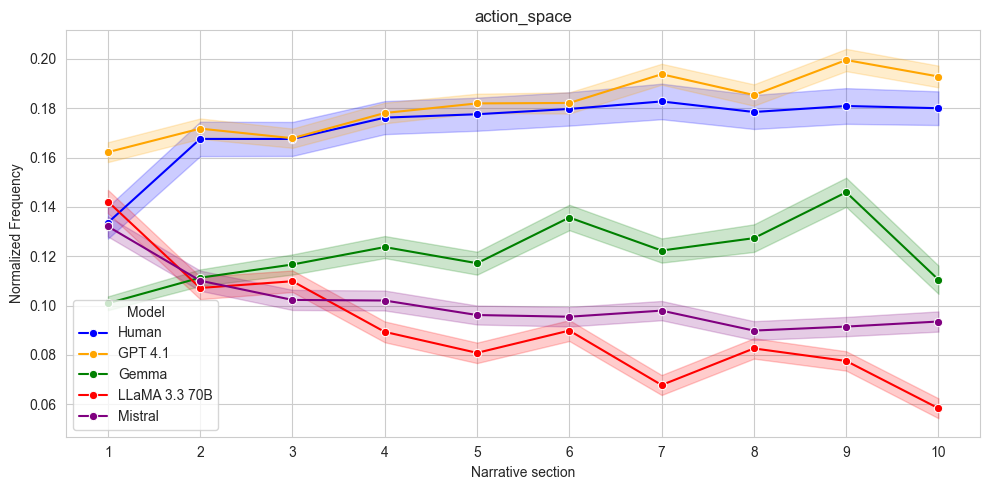}
    \end{subfigure}
    \hfill
    \begin{subfigure}[t]{0.48\linewidth}
        \includegraphics[width=\linewidth]{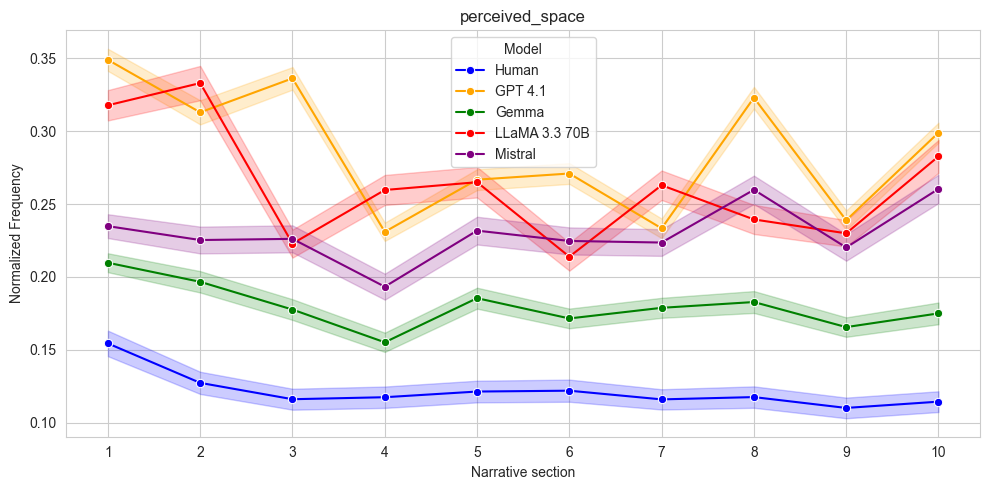}
    \end{subfigure}
    \vspace{0.5em}
    \begin{subfigure}[t]{0.48\linewidth}
        \includegraphics[width=\linewidth]{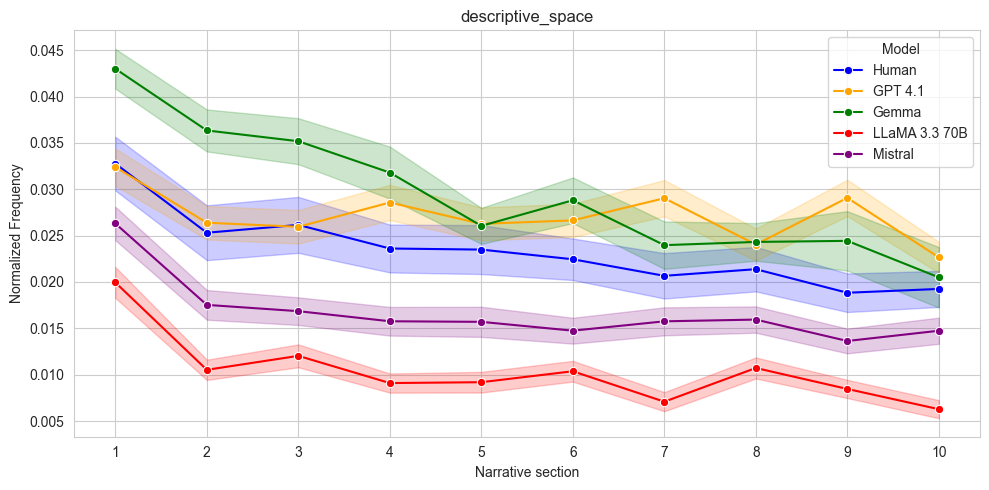}
    \end{subfigure}
    \hfill
    \begin{subfigure}[t]{0.48\linewidth}
        \includegraphics[width=\linewidth]{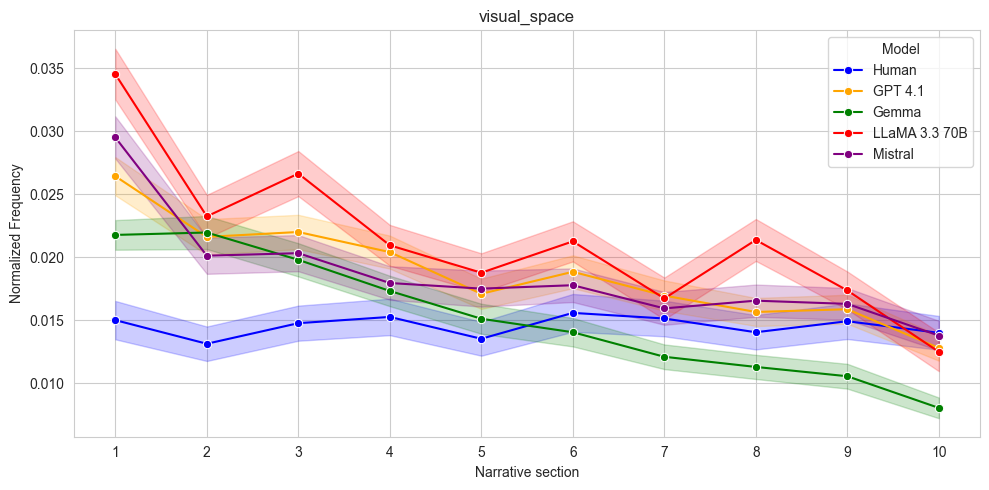}
    \end{subfigure}
    \vspace{0.5em}
    \begin{subfigure}[t]{\linewidth}
        \includegraphics[width=\linewidth]{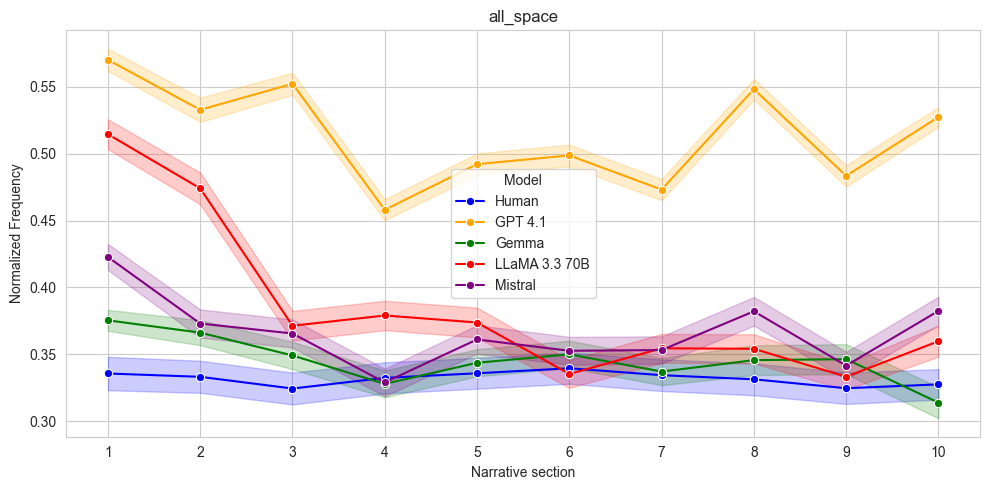}
    \end{subfigure}
\end{minipage}
\hfill
\begin{minipage}[t]{0.48\textwidth}
    \centering
    \begin{subfigure}[t]{0.48\linewidth}
        \includegraphics[width=\linewidth]{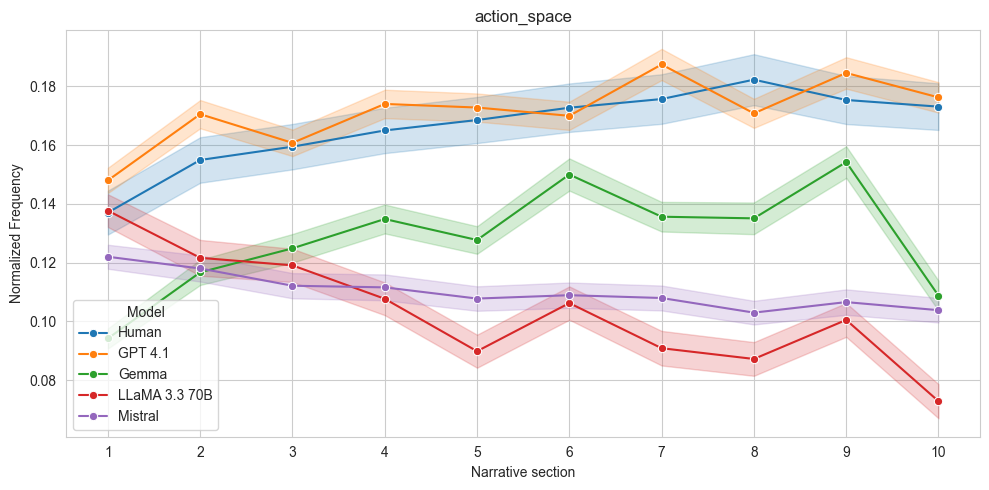}
    \end{subfigure}
    \hfill
    \begin{subfigure}[t]{0.48\linewidth}
        \includegraphics[width=\linewidth]{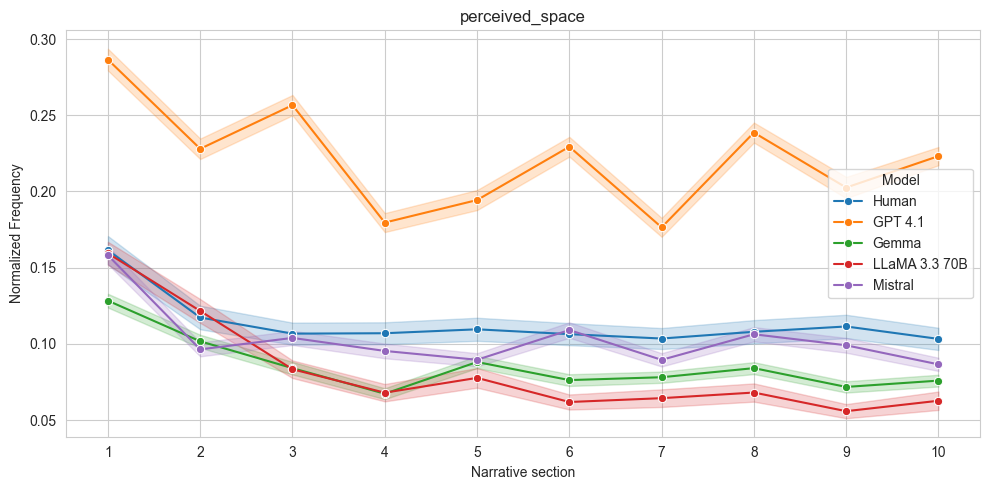}
    \end{subfigure}
    \vspace{0.5em}
    \begin{subfigure}[t]{0.48\linewidth}
        \includegraphics[width=\linewidth]{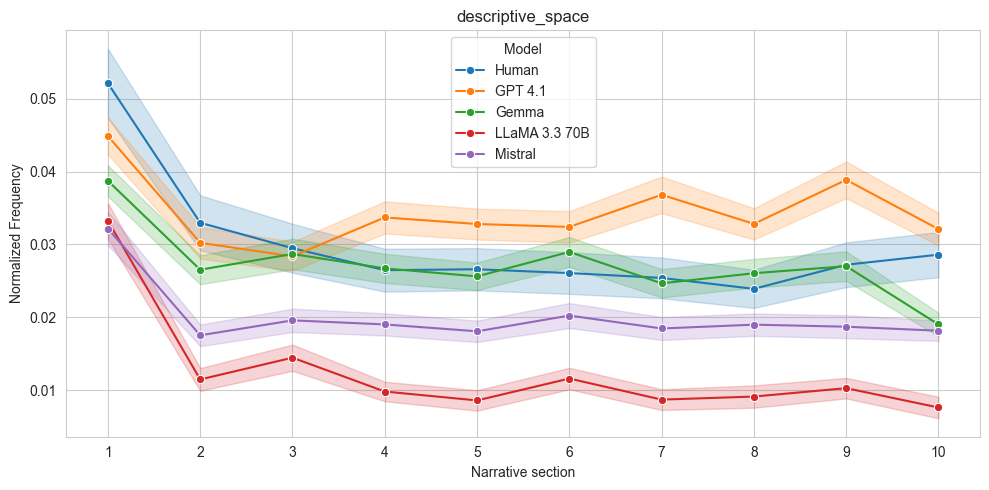}
    \end{subfigure}
    \hfill
    \begin{subfigure}[t]{0.48\linewidth}
        \includegraphics[width=\linewidth]{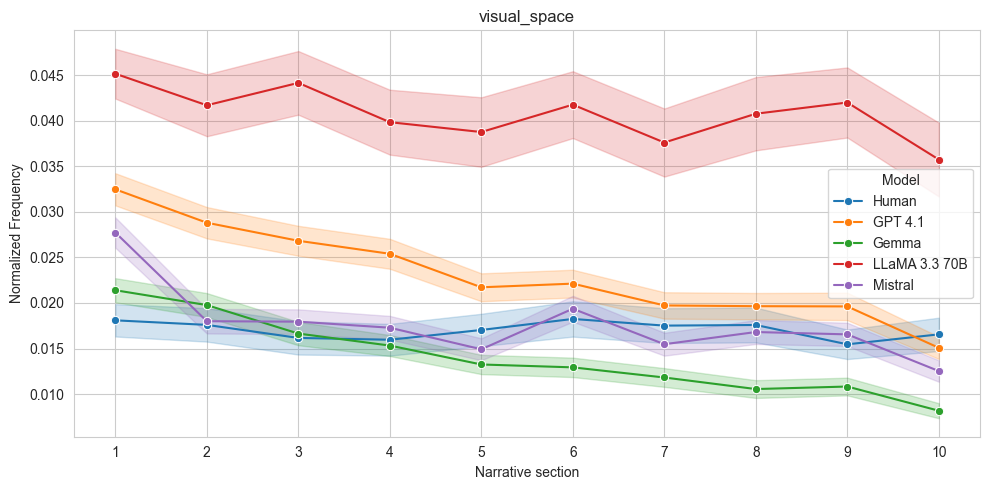}
    \end{subfigure}
    \vspace{0.5em}
    \begin{subfigure}[t]{\linewidth}
        \includegraphics[width=\linewidth]{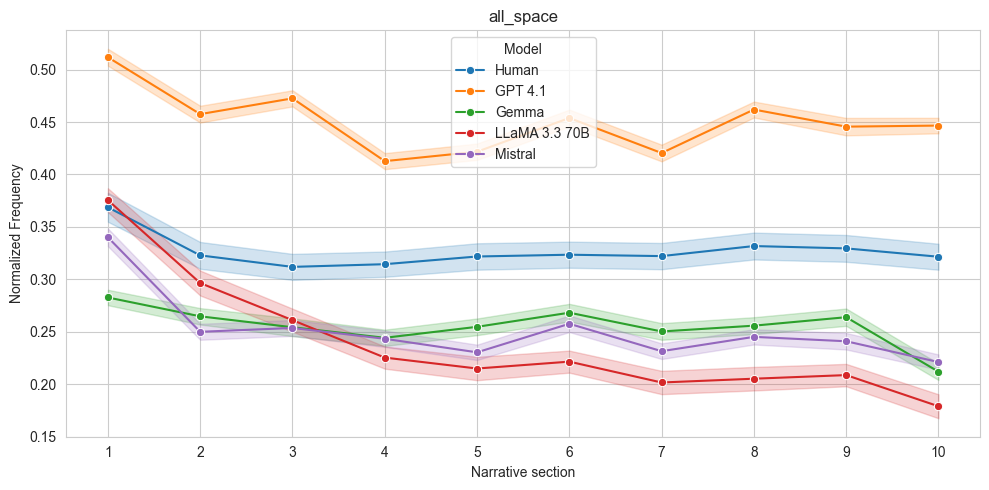}
    \end{subfigure}
\end{minipage}
\caption{Temporal distribution of setting categories across narrative 
time for English (left) and German (right).}
\label{fig:five-boxes}
\end{figure*}

Figure~\ref{fig:five-boxes} shows the full temporal distribution 
of setting categories across narrative time for English and German.

\section{Model Identifiability from Spatial Distributions}
\label{app:classifier}

\begin{figure*}[h]
    \centering
    \begin{subfigure}[t]{0.48\textwidth}
        \includegraphics[width=\linewidth]{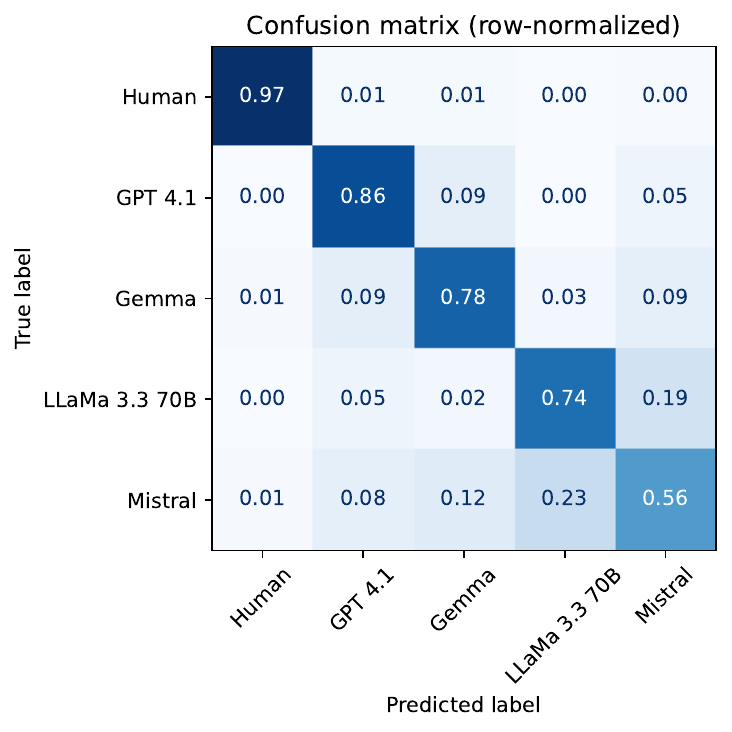}
        \caption{English}
        \label{fig:confusion-en}
    \end{subfigure}
    \hfill
    \begin{subfigure}[t]{0.48\textwidth}
        \includegraphics[width=\linewidth]{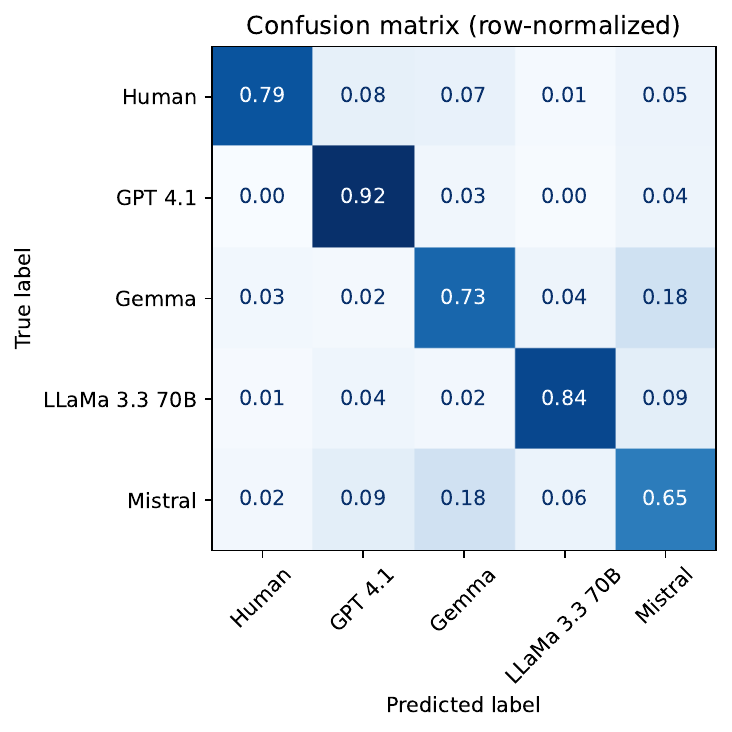}
        \caption{German}
        \label{fig:confusion-de}
    \end{subfigure}
    \caption{Row-normalized confusion matrices for model 
    identification from spatial category distributions.}
    \label{fig:confusion-matrices}
\end{figure*}

A random forest classifier trained on per-section spatial category proportions identified the source model well above chance in both languages (confusion matrices in Figures~\ref{fig:confusion-en} and~\ref{fig:confusion-de}). In English, human-authored texts are most reliably identified (recall = 0.97), while Mistral~3.2 is hardest (0.56). The main source of confusion is between LlaMA~3.3 and Mistral~3.2, consistent with their closely aligned setting distributions. In German, human recall drops to 0.79 as LLM distributions converge toward the baseline.

\section{Temporal Stability of Human-Authored Corpora}
\label{app:temporal}

\begin{figure*}[t]
    \centering
    \begin{subfigure}[t]{0.48\textwidth}
        \centering
        \includegraphics[width=\textwidth]{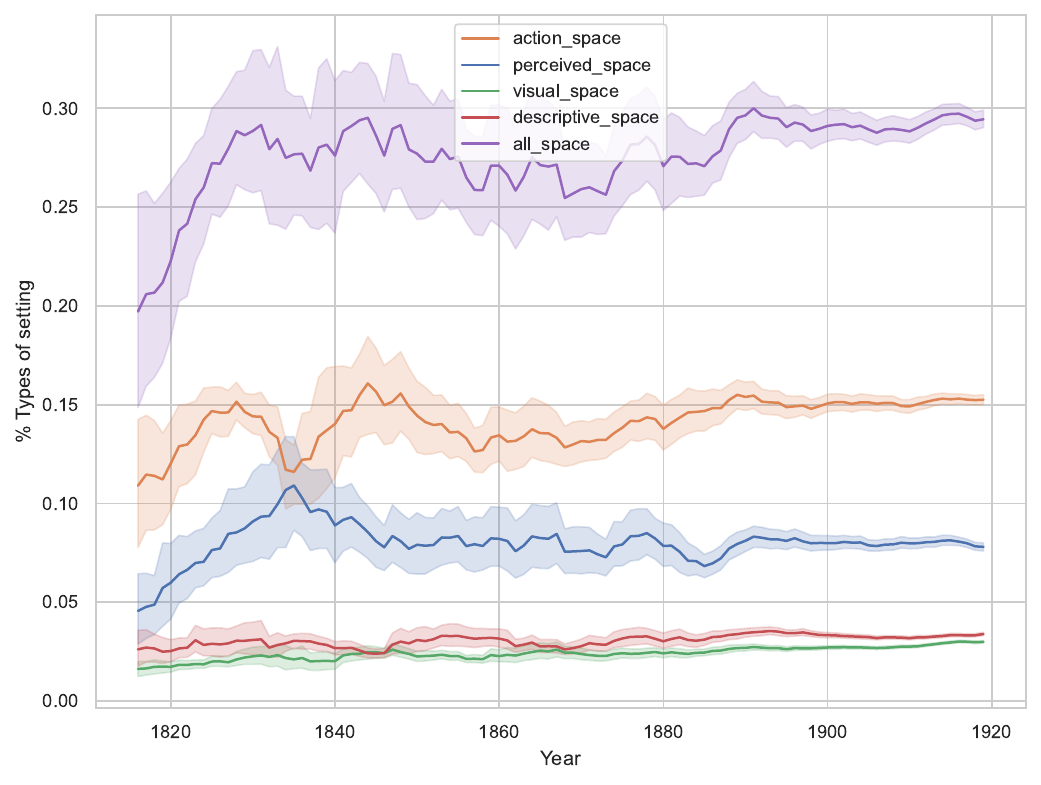}
        \caption{English corpus (1800--1920).}
        \label{fig:hist-time-eng}
    \end{subfigure}
    \hfill
    \begin{subfigure}[t]{0.48\textwidth}
        \centering
        \includegraphics[width=\textwidth]{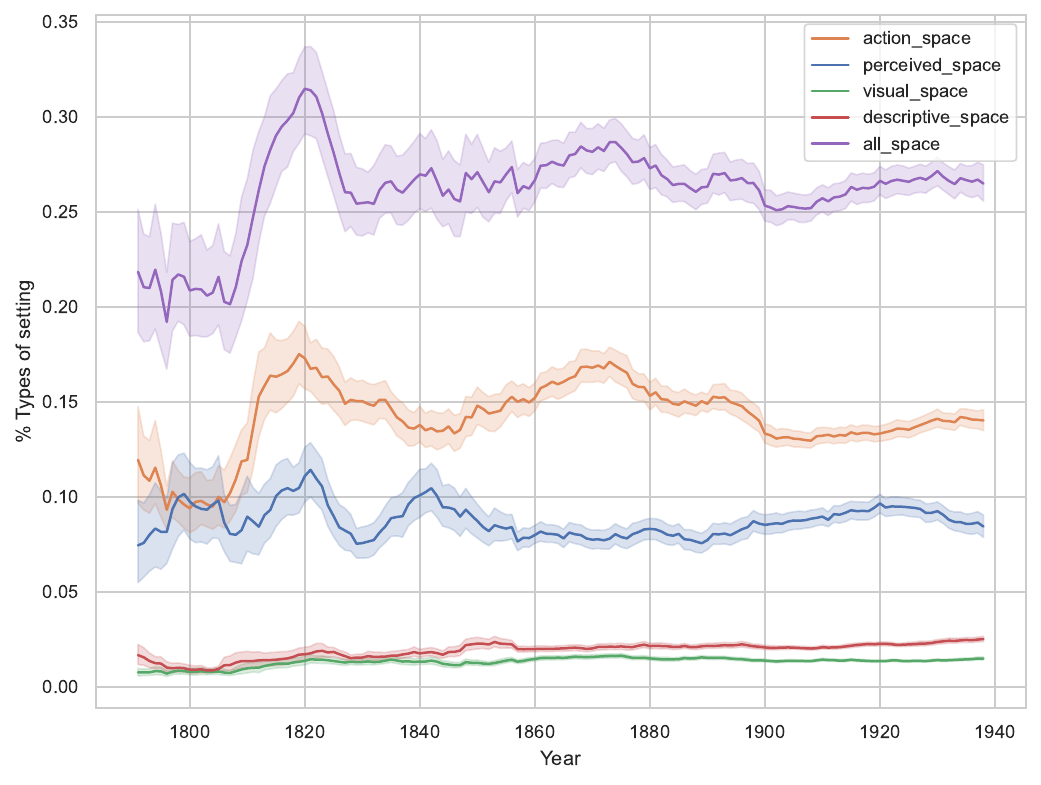}
        \caption{German corpus (1780--1940), reproduced from 
        \citet{author_forthcoming}.}
        \label{fig:hist-time-ger}
    \end{subfigure}
    \caption{Normalized frequencies of setting categories 
    across publication years in the human-authored corpora. 
    Shaded bands indicate 95\% confidence intervals. 
    Wider intervals in earlier decades reflect smaller 
    sample sizes. Distributions are broadly stable across 
    the main period of each corpus.}
    \label{fig:hist-time}
\end{figure*}

Figure~\ref{fig:hist-time} shows the distribution of 
setting categories across publication years in the 
human-authored corpora for English and German. 
Distributions are broadly stable across the main period 
of each corpus, supporting the assumption that observed 
differences between human-authored and AI-generated 
texts are not driven by historical sub-period.

\section{Full GLMM Results}
\label{app:glmm}

Table~\ref{tab:glmm-genre} reports the full Type III Wald $\chi^2$ tests for the GLMM described in Section~\ref{sec:stats}, fitted separately
for each setting category and language. These are the models from which all
test statistics reported in Section~\ref{sec:temporal} are taken. Genre is
coded as novel vs.\ other fiction and, as noted in Section~\ref{sec:stats}, is
matched across author conditions by construction. 

\begin{table*}[t]
\centering
\small
\begin{tabular}{llrrrr}
\toprule
& \textbf{Category} & \textbf{model} & \textbf{section} & \textbf{genre} & \textbf{model\,$\times$\,section} \\
& & $\chi^2(4)$ & $\chi^2(9)$ & $\chi^2(1)$ & $\chi^2(36)$ \\
\midrule
\multirow{5}{*}{\rotatebox{90}{English}}
& Perceived space   & 2509.37\rlap{$^{***}$} &   10.25\rlap{$^{\text{n.s.}}$} &  18.63\rlap{$^{***}$} & 2013.48\rlap{$^{***}$} \\
& Action space      &  456.78\rlap{$^{***}$} &   49.05\rlap{$^{***}$}         &  23.14\rlap{$^{***}$} & 2660.88\rlap{$^{***}$} \\
& Visual space      &  273.88\rlap{$^{***}$} &   49.19\rlap{$^{***}$}         &   0.39\rlap{$^{\text{n.s.}}$} &  457.07\rlap{$^{***}$} \\
& Descriptive space &  259.41\rlap{$^{***}$} &  537.06\rlap{$^{***}$}         &   2.41\rlap{$^{\text{n.s.}}$} &  487.61\rlap{$^{***}$} \\
& All space         & 1834.64\rlap{$^{***}$} &   14.24\rlap{$^{\text{n.s.}}$} &   0.41\rlap{$^{\text{n.s.}}$} & 1889.35\rlap{$^{***}$} \\
\midrule
\multirow{5}{*}{\rotatebox{90}{German}}
& Perceived space   & 1978.04\rlap{$^{***}$} &   81.78\rlap{$^{***}$} &  88.51\rlap{$^{***}$}         & 1860.15\rlap{$^{***}$} \\
& Action space      &  307.42\rlap{$^{***}$} &   54.39\rlap{$^{***}$} &  14.86\rlap{$^{***}$}         & 1449.79\rlap{$^{***}$} \\
& Visual space      &  700.58\rlap{$^{***}$} &   21.39\rlap{$^{*}$}   &  26.44\rlap{$^{***}$}         &  476.03\rlap{$^{***}$} \\
& Descriptive space &  243.01\rlap{$^{***}$} &  242.25\rlap{$^{***}$} &   1.05\rlap{$^{\text{n.s.}}$} &  738.40\rlap{$^{***}$} \\
& All space         & 1478.24\rlap{$^{***}$} &   18.75\rlap{$^{*}$}   &  11.51\rlap{$^{***}$}         & 1749.91\rlap{$^{***}$} \\
\bottomrule
\end{tabular}
\caption{Type III Wald $\chi^2$ tests for the GLMM,
\texttt{space \textasciitilde{} author $\times$ section + genre + (1 | story)},
fitted separately per setting category and language.
$^{***}p<.001$, $^{*}p<.05$, n.s. $=$ not significant.}
\label{tab:glmm-genre}
\end{table*}

\section{Chapter-Boundary Position}
\label{app:chapters}

\begin{figure*}[ht]
    \centering
    \includegraphics[width=\textwidth]{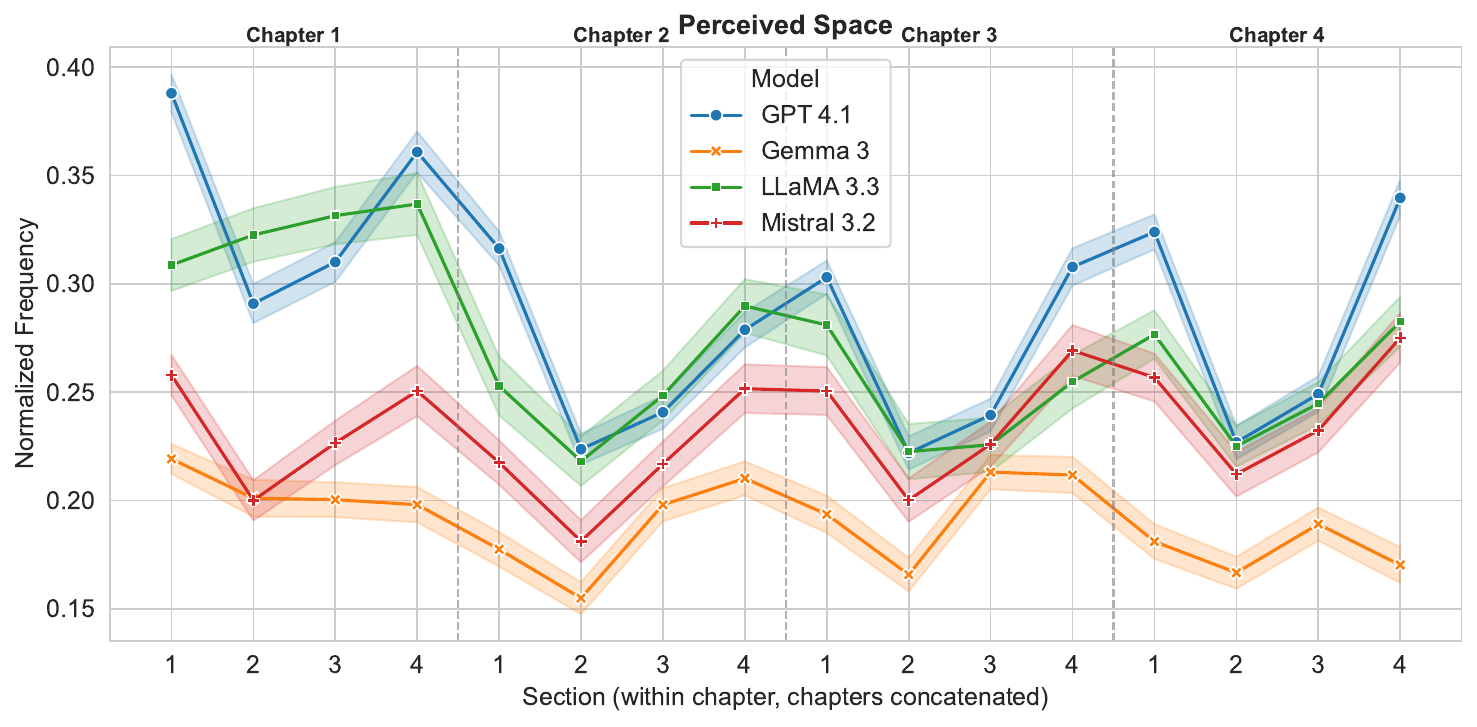}
    \caption{Normalized frequency of perceived space by position within
    chapter, English. Each chapter is divided into four equal-length sections
    and the four chapters are shown consecutively. The dashed line indicates
    the human baseline.}
    \label{fig:chapter-en}
\end{figure*}

\begin{figure*}[ht]
    \centering
    \includegraphics[width=\textwidth]{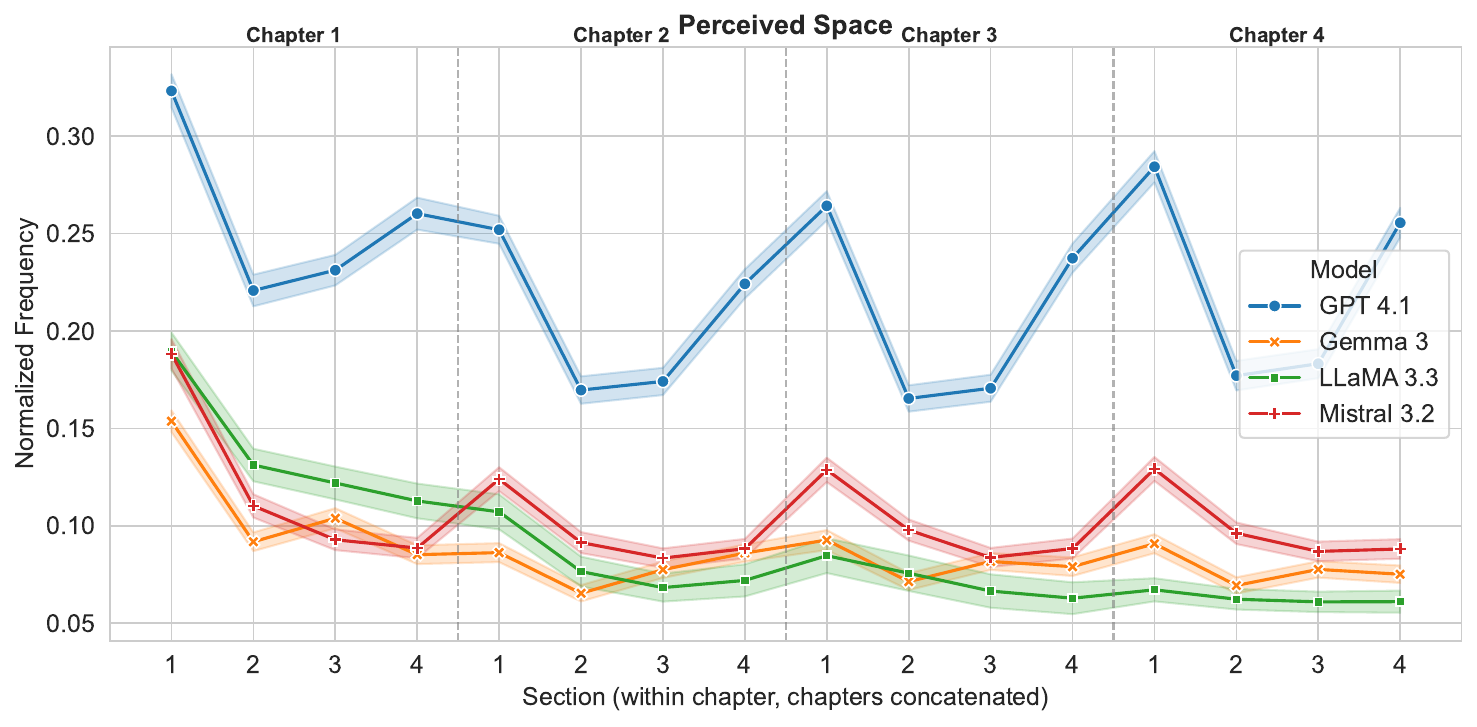}
    \caption{Normalized frequency of perceived space by position within
    chapter, German. Each chapter is divided into four equal-length sections
    and the four chapters are shown consecutively. The dashed line indicates
    the human baseline.}
    \label{fig:chapter-de}
\end{figure*}

Figures~\ref{fig:chapter-en} and~\ref{fig:chapter-de} show the profiles for perceived space by position within chapter, for English and German respectively, and Table~\ref{tab:chapter-means} reports the corresponding means. 

Beyond perceived space, action space shows the mirror pattern, dropping at chapter boundaries relative to chapter middles, while visual and descriptive space decline slightly and steadily across each chapter, with a total spread of at most 0.006 and no boundary peak. The boundary effect is thus specific to perceived and, inversely, action space.

\begin{table}[ht]
\centering
\small
\begin{tabular}{lcccc}
\toprule
\textbf{Model} & \textbf{Q1} & \textbf{Q2} & \textbf{Q3} & \textbf{Q4} \\
\midrule
\multicolumn{5}{l}{\textit{English}} \\
GPT~4.1      & 0.333 & 0.241 & 0.260 & 0.322 \\
LlaMA~3.3    & 0.280 & 0.247 & 0.263 & 0.291 \\
Mistral~3.2  & 0.246 & 0.198 & 0.225 & 0.262 \\
Gemma~3      & 0.193 & 0.172 & 0.200 & 0.197 \\
Human        & \multicolumn{4}{c}{0.083} \\
\midrule
\multicolumn{5}{l}{\textit{German}} \\
GPT~4.1      & 0.281 & 0.183 & 0.190 & 0.244 \\
Mistral~3.2  & 0.143 & 0.099 & 0.087 & 0.088 \\
LlaMA~3.3    & 0.113 & 0.087 & 0.080 & 0.077 \\
Gemma~3      & 0.106 & 0.074 & 0.085 & 0.081 \\
Human        & \multicolumn{4}{c}{0.084} \\
\bottomrule
\end{tabular}
\caption{Normalized frequency of perceived space by quarter within chapter (Q1--Q4), averaged over the four chapters. Human-authored texts are not chapter-segmented; their corpus mean is given as a single reference value.}
\label{tab:chapter-means}
\end{table}

\section{Generation Length}
\label{app:length}

We generated between 2,500 and 3,000 tokens per chapter with an enforced maximum of 3,000 tokens, stopping when the model emitted an end-of-sequence token or reached the cap, for four chapters and a total of 10,000--12,000 tokens per story. Where a model reproduced the seeding sentence verbatim, we removed the duplicated first sentence; where generation stopped mid-sentence, we removed the trailing incomplete sentence. Table~\ref{tab:length} reports the resulting story lengths in words.

\begin{table}[ht]
\centering
\small
\begin{tabular}{lrrrrr}
\toprule
\textbf{Model} & \textbf{Mean} & \textbf{SD} & \textbf{Min} & \textbf{P10} & \textbf{Max} \\
\midrule
\multicolumn{6}{l}{\textit{English}} \\
GPT~4.1     & 8,456 & 324 & 6,971 & 8,035 &  9,317 \\
Gemma~3     & 8,525 & 540 & 5,472 & 7,860 &  9,801 \\
LlaMA~3.3   & 9,376 & 356 & 8,120 & 8,898 & 10,242 \\
Mistral~3.2 & 9,353 & 452 & 7,479 & 8,766 & 10,526 \\
\midrule
\multicolumn{6}{l}{\textit{German}} \\
GPT~4.1     & 6,522 & 807 & 2,233 & 5,815 & 7,748 \\
Gemma~3     & 7,074 & 382 & 5,518 & 6,572 & 8,438 \\
LlaMA~3.3   & 6,788 & 260 & 5,899 & 6,457 & 7,628 \\
Mistral~3.2 & 7,255 & 515 & 3,723 & 6,568 & 8,715 \\
\bottomrule
\end{tabular}
\caption{Story length in words per model and language ($n = 1{,}000$ each).}
\label{tab:length}
\end{table}

\end{document}